\documentclass{article} 
\usepackage{iclr2026_conference,times}
\usepackage[numbers]{natbib}

\usepackage{amsmath,amsfonts,bm}

\def\eqref#1{equation~\ref{#1}}

\def\1{\bm{1}}

\DeclareMathAlphabet{\mathsfit}{\encodingdefault}{\sfdefault}{m}{sl}
\SetMathAlphabet{\mathsfit}{bold}{\encodingdefault}{\sfdefault}{bx}{n}

\newcommand{\spacing}[1]{\vspace{#1}}
\newcommand{\valstd}[2]{$#1$ {\tiny $\pm #2$}}

\newcommand{\ours}{{{CDGS}}}

\usepackage{wrapfig}
\usepackage{hyperref}
\usepackage{url}
\usepackage{multirow}

\usepackage{makecell}
\usepackage{graphicx}
\usepackage{epsfig} 
\usepackage{mathptmx}
\usepackage{amsmath} 

\usepackage{amssymb}  
\usepackage{amsthm}
\usepackage{wasysym}
\usepackage{mathrsfs}
\usepackage[mathcal]{euscript}
\usepackage{bbm}
\usepackage{color}
\usepackage{lipsum}
\usepackage{algorithm}
\usepackage{algpseudocode}
\usepackage{tcolorbox}
\tcbuselibrary{breakable}
\usepackage[normalem]{ulem}
\usepackage{tabularx}
\usepackage{booktabs}
\usepackage{array}

\usepackage{subcaption}
\usepackage{graphicx}

\newenvironment{rebuttal}
  {\medskip \noindent}
  {\medskip}

\newcommand{\changes}[1]{{{{#1}}}}

\title{Compositional Diffusion with Guided Search for Long-Horizon Planning}

\author{Utkarsh A.~Mishra, David He, Yongxin Chen and Danfei Xu \\
Georgia Institute of Technology \\
\texttt{utkarshm.robo@gatech.edu}
}

\iclrfinalcopy  
\begin{document}

\maketitle

\begin{abstract}


Generative models have emerged as powerful tools for planning, with compositional approaches offering particular promise for modeling long-horizon task distributions by composing together local, modular generative models. This compositional paradigm spans diverse domains, from multi-step manipulation planning to panoramic image synthesis to long video generation. However, compositional generative models face a critical challenge: when local distributions are multimodal, existing composition methods average incompatible modes, producing plans that are neither locally feasible nor globally coherent. We propose Compositional Diffusion with Guided Search (\ours{}), which addresses this \emph{mode averaging} problem by embedding search directly within the diffusion denoising process. Our method explores diverse combinations of local modes through population-based sampling, enforces global consistency through iterative resampling between overlapping segments, and prunes infeasible candidates using likelihood-based filtering. \ours{} matches oracle performance on seven robot manipulation tasks, outperforming baselines that lack compositionality or require long-horizon training data. The approach generalizes across domains, enabling coherent text-guided panoramic images and long videos through effective local-to-global message passing. More details: \url{https://cdgsearch.github.io/}

\end{abstract}

\begin{wrapfigure}{r}{0.55\columnwidth}  
    \centering
    \vspace{-3.5em}
    \includegraphics[width=0.46\columnwidth,trim={0cm 0.8cm 0cm 0.5cm}]{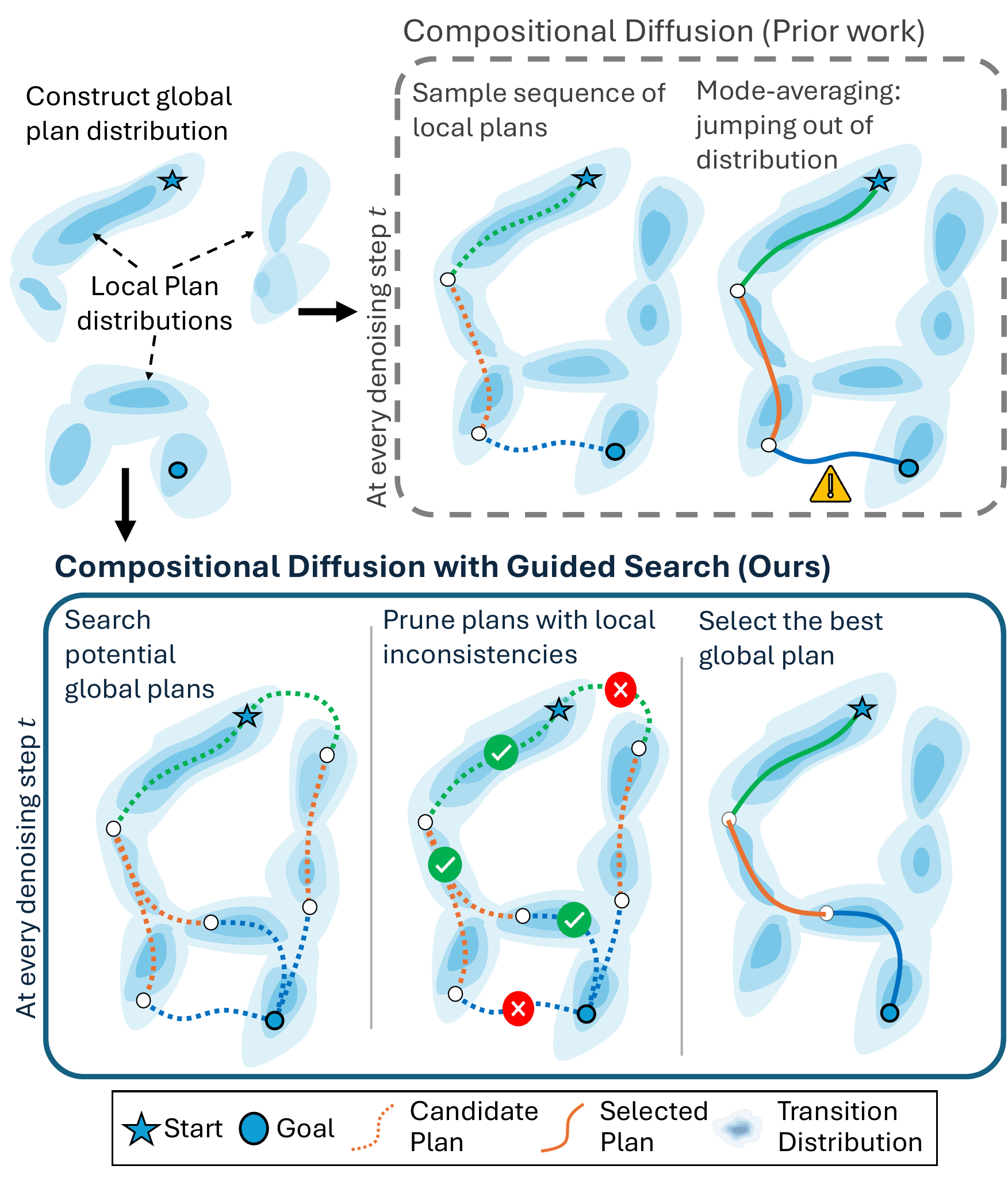}
    \caption{\textbf{Compositional Diffusion with Guided Search (CDGS)} composes short-horizon plan distributions to sample long-horizon goal-directed plans directly at inference. Unlike na\"ive compositional sampling, it explores diverse plans and filters locally inconsistent paths to avoid ``mode averaging", yielding globally coherent plans.
    }
    \label{fig:teaser}
    \vspace{-2em}
\end{wrapfigure}


\section{Introduction}
\label{sec:introduction}

Synthesizing coherent long sequences is a crucial and challenging task, requiring reasoning over extended horizons. This task arises naturally in various domains: robotic actions must enable future steps, parts of a panorama must align semantically, and subjects in a video must remain consistent across hundreds of frames.

\begin{figure}[t]
    \centering
    \includegraphics[width=0.9\linewidth, trim={0cm 0.5cm 0.5cm 0.5cm}]{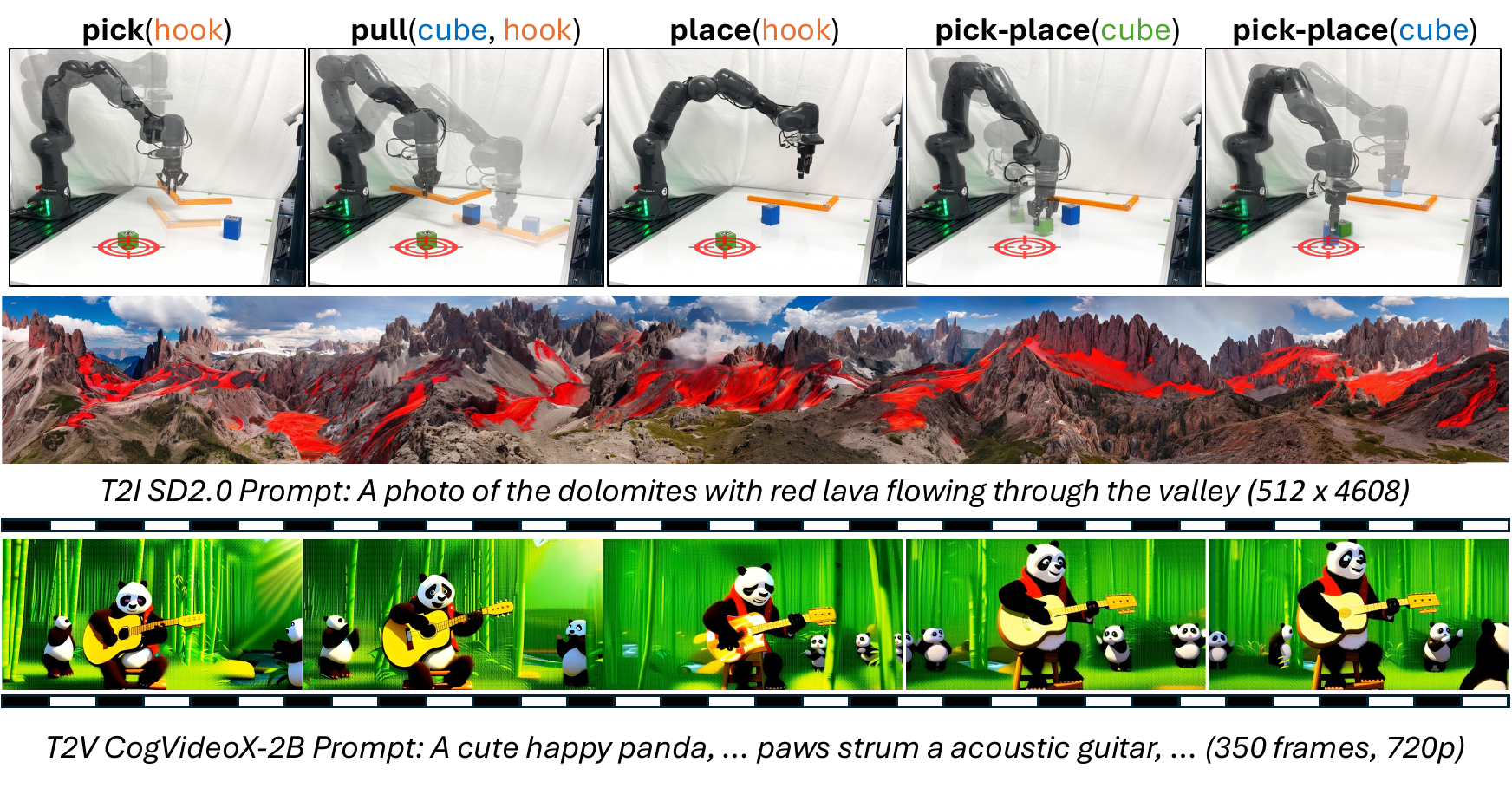}
    \caption{
    \textbf{Applications of \ours{}.}  (Top) Long horizon motion planning: \ours{} discovers a valid multi-step plan to move the blue cube to the green cube’s original position via : (1) using the {hook} to pull {blue cube} in workspace, (2) displace the {green cube} to make space and (3) moving the {blue cube} to the target position. (Mid) \ours{} generates coherent panoramic images. (Bottom) \ours{} can stitch short clips to generate consistent, longer videos.
    }
    \label{fig:applications}
    \spacing{-1em}
\end{figure}

Recent work leverages generative models to learn long sequence distributions~\cite{janner22diffuser, ajay2023decisiondiffuser}, with diffusion models~\cite{sohl-dickstein_deep_2015, ho2020denoising} gaining popularity for modeling multi-modal data~\cite{dhariwal2021diffusion, ho2022cfg}. However, full-sequence data is expensive to acquire, and monolithic models 
fail to generalize beyond training horizons~\cite{du2024position}. As an alternative, compositional generation effectively combines short-horizon local 
distributions 
to sample long-horizon global plans~\cite{zhang2023diffcollage, mishra2023gsc, luo2025generative}—e.g., chaining skills for task planning, connecting images into panoramas, or stitching clips into videos. While this improves data-efficiency and allows extrapolation beyond training data, it introduces a \textbf{critical challenge}: as local plan distributions become highly \emph{multimodal}, the 
distribution
of global plans inherits combinatorial multi-modality. For example, in the robotics scenario in~\autoref{fig:applications}, because the robot has a large combination of actions and objects it can act on, the search space of possible plans grows exponentially with the length of the planning horizon.

Existing methods for compositional generation offer a promising approach, using \emph{score-averaging} to compose modes of local distributions into a global
distribution~\cite{zhang2023diffcollage, mishra2023gsc}. However, these methods have an \textbf{important limitation}: their inability to handle the combinatorial multi-modality leads them to \emph{average} incompatible local modes~(\emph{mode-averaging}), ultimately producing invalid global plans. Addressing such complex multi-modal distributions requires inference methods that jointly reason about compatibility between local modes and effectively navigate the exponentially large search space.

To address the challenge and overcome the limitation, we aim to identify compatible sequences of local modes that compose into a globally coherent plan. Given the diversity and multi-modality of the search space, we take inspiration from classical search techniques and introduce Compositional Diffusion with Guided Search~(\ours{}), a guided search mechanism integrated into the diffusion denoising process as illustrated in~\autoref{fig:teaser}. To facilitate the search during inference, at each diffusion timestep, our method introduces two key components: (i) \textbf{iterative resampling} to enhance local-global message passing in compositional diffusion to propose
globally plausible candidates,
and (ii) \textbf{likelihood-based pruning} to remove incoherent candidates that fall into low-likelihood regions due to 
mode-averaging. Together, these components enable 
\ours{} to
efficiently sample coherent long-horizon plans. For robotics tasks, our method outperforms or is on par with baselines that 
lack compositionality 
or use long-horizon data for training, respectively. We also show the efficacy of our method in long text-to-image and text-to-video tasks~(\autoref{fig:applications}), producing more coherent and consistent generations. 
\section{Background}
\label{sec:background}

\textbf{Problem formulation.} A long-horizon plan generation problem is characterized by the task of constructing a global plan $\tau = (x_1, \ldots, x_N)$ as a sequence of variables $x_i$, by sampling from the joint distribution $p(\tau)$. The problem becomes goal-directed if $\tau$ must connect a given start $x_1 = x_{s}$ to a desired goal $x_N = x_{g}$.
Such problems arise in diverse domains: long-horizon manipulation planning, panoramic images, and long videos. While modeling the full joint distribution $p(\tau)$ would directly model all dependencies between any $x_i$, it usually entails end-to-end learning from long-horizon data, which can be infeasible or expensive to obtain. 
In the absence of long-horizon data, a promising strategy is to approximate the joint distribution $p(\tau)$ with a factor graph of overlapping local distributions that can be learned from short-horizon data. For the joint variable $\tau = (x_1, x_2, \ldots, x_N)$, we construct a factor graph \cite{koller2009probabilistic} connecting variable nodes $\{x_i\}_{i=1}^{N}$ and factor nodes $\{y_j\}_{j=1}^{M}$, where each factor $y_j$ represents the joint distribution of contiguous subsequences of $\tau$. For example, we represent $\tau = (x_1, x_2, \ldots, x_5)$ with factors $y_1 = (x_1, x_2, x_3), y_2 = (x_3, x_4, x_5)$.
With this, we construct the joint distribution $p(\tau)$ using the Bethe approximation \cite{yedidia2005bethe}:
\begin{align}
\label{eq:bethe}
\small
    p(\tau) &:= \frac{p(x_1, x_2, x_3) p(x_3, x_4, x_5)\ldots}{p(x_3)\ldots}
    = \frac{\prod_{j=1}^{M} p(y_j)}{\prod_{i=1}^{N} p(x_i)^{d_i - 1}}\,
\end{align}
where $d_i$ is the degree of the variable node $x_i$. 
This representation enables sampling from the long-horizon distribution $p(\tau)$ using only samples drawn from a short-horizon distribution $p(y)$.

\textbf{Diffusion models.} Diffusion models are defined by a forward process that progressively injects noise into the data distribution $p(y^{(0)})$ and a reverse diffusion process that iteratively removes the noise by approximating $\nabla \log p$ to recover the original data distribution. For a given noise injection schedule $\alpha_t$, forward noising adds a Gaussian noise $\epsilon$ to clean samples s.t. $y^{(t)} = \sqrt{\alpha_t}y^{(0)} + \sqrt{1-\alpha_t} \epsilon$. With $p_t$ being the distribution of noisy samples, the denoising is performed using the score function $\nabla_{y^{(t)}} \log p_t(y^{(t)})$ often estimated by a neural network $\epsilon_\theta(y^{(t)}, t)$ learned via minimizing the score matching loss~\cite{hyvarinen2005scorematching} given by $\mathbb{E}_{t,y^{(0)}}[|\epsilon - \epsilon_\theta(y^{(t)}, t)|^2]$. Such a score function allows denoising the noise samples via sampling from
\begin{equation}\small \label{eq:tweedie}
p(y^{(t-1)}|y^{(t)}) = \mathcal{N}\left(y^{(t-1)}; \sqrt{\alpha_{t-1}}\hat{y}_0^{(t)} + \sqrt{1-\alpha_{t-1} - \sigma_t^2}  \epsilon(y^{(t)},t), \sigma_t^2\mathbf{I}\right)
\end{equation}
where $\hat{y}_0^{(t)} = \frac{y^{(t)} - \sqrt{1-\alpha_t} \epsilon_\theta(y^{(t)},t)}{\sqrt{\alpha_t}}$ is the Tweedie estimate of the clean sample distribution at denoising step $t$ and $\sigma_t$ controls stochasticity~\cite{song2020ddim}. Several works have leveraged the flexibility of the denoising process in performing post-hoc guidance~\cite{ho2022cfg} and plug-and-play generation~\cite{liu2022compositional, du2023reduce}.

\begin{figure}[t]
    \centering
    \includegraphics[width=\linewidth, trim={0cm 0.2cm 0cm 0.5cm}]{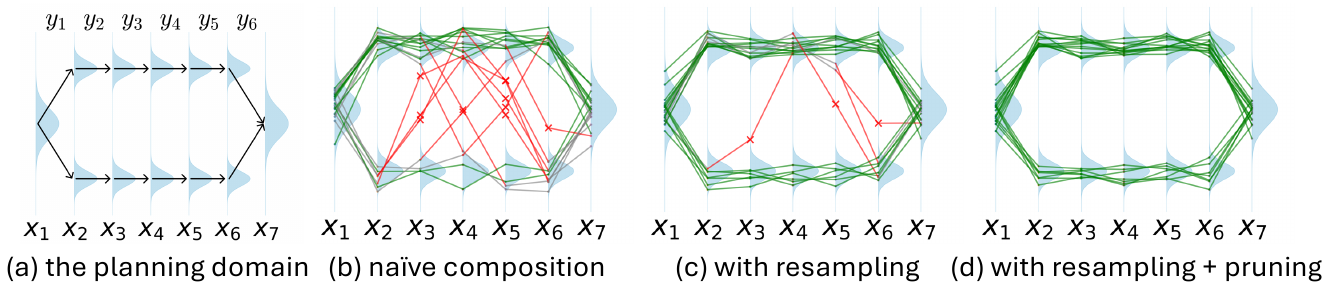}
    \caption{\textbf{Running example.} (a) Consider a 1D-domain of $\{x_{1:7}\}$ variable distributions and $\{y_{1:6}\}$ feasible directed transitions between the variables. There are two feasible long-horizon plans from start~($x_1$) to goal~($x_7$): one through the top and one through the bottom. 
    (b) in naive-composition, sampled plans may choose to start in the top and end at the bottom, or vice versa. When this happens, the intermediate models $\{y_{2:5}\}$ will average the modes of intermediate variables $\{x_{2:6}\}$ to satisfy both constraints, manifesting in infeasible transitions (red) (c) adding \textbf{iterative resampling} reduces the frequency of mode-averaging (d) adding \textbf{pruning} eliminates plans with infeasible $y$ 
    }
    \label{fig:multimodality}
    \vspace{-1em}
\end{figure}

\textbf{Compositional sampling with diffusion models}. Under the diffusion model formulation, 
we can \emph{compositionally} sample \cite{du2020compositional, zhang2023diffcollage} from the factor graph representation of $p(\tau)$ by calculating the score $\nabla \log p(\tau)$ as a sum of factor and variable scores following~\autoref{eq:bethe}:
\everymath{\displaystyle} 
\begin{align}
\label{eq:compositional_score}\small
\nabla\log p(\tau) := \sum_{j=1}^{M} \nabla\log p(y_j) + \sum_{i=1}^{N}(1 - d_i) \nabla\log p(x_i) 
\end{align}

In practice, our factor graph is a chain, so overlapping variables (i.e., the ones shared between neighboring factors $y_j$ and $y_{j+1}$) have degree~$d_i=2$ while non-overlapping ones have $d_i=1$~(i.e. their marginals have no contribution to $\nabla\log p(\tau)$). For overlapping variables, we approximate the marginal scores using the average of the conditional score: {\small$\nabla \log p(x_i) \approx \frac{1}{2}\Big[\nabla \log p_{y_j}(x_i | \ldots) +\nabla \log p_{y_{j+1}}(x_i | ...)\Big]$} where $x_i\in y_j \cap y_{j+1}$ denotes the overlapping variable between $y_j$ and $y_{j+1}$. This, along with the scores of $p(y_j)$ computed from local distribution, allows us to formulate the global compositional score $\nabla \log p(\tau)$ using~\autoref{eq:compositional_score}.  While this formulation enables generalization beyond the lengths seen during training, it comes with limitations described in~\autoref{sec:method}.

\section{Method}
\label{sec:method}

\textbf{Challenge: Compositional sampling with multi-modal distributions.} Solving long-horizon tasks requires constructing a coherent global plan distribution 
that induces an exponentially large search space and requires reasoning about long-horizon dependencies. 
 Data scarcity prohibits directly learning the target global plan distribution~$p(\tau)$, so a convincing alternative is to approximate it as a composition of local plan distribution $p(y)$~(using~\autoref{eq:bethe}). Thus, one can sample short-horizon local plans $y_{1:M}\sim p(y)$ and compose them with suitable overlaps to form a coherent $\tau$. However, as the diversity of feasible local behaviors increases, $p(y)$ becomes highly multi-modal and composing such distributions causes $p(\tau)$ to inherit combinatorial multi-modality—where each mode of the global plan distribution corresponds to a distinct sequence of modes from the local plan distribution. In this setting, na\"ive compositional methods~(\cite{mishra2023generative}) 
that merge distributions $y_{1:M}\sim p(y)$ via score averaging~
(\autoref{eq:compositional_score}) 
often fail due to the mode-averaging issue: selecting high-likelihood local segments that, while individually plausible, result in incompatible mode sequences—leading to inconsistent overlaps and incoherent global plans. 
A natural way to address multi-modality is to explore diverse modes during 
sampling,
an idea recently explored by inference-time scaling approaches~\cite{ma2025inference, zhang2025inference}.
However, these methods are limited to sampling from standalone distributions and not 
a composed sequence of distributions. The key challenge is to generate a feasible sequence of local plans that collectively form a coherent global plan—requiring a
sampling algorithm that reasons over structured combinations of modes rather than collapsing into incoherent averages.

\begin{figure}[t]
    \centering
    \includegraphics[width=\linewidth, trim={0cm 5.5cm 0cm 0cm}]{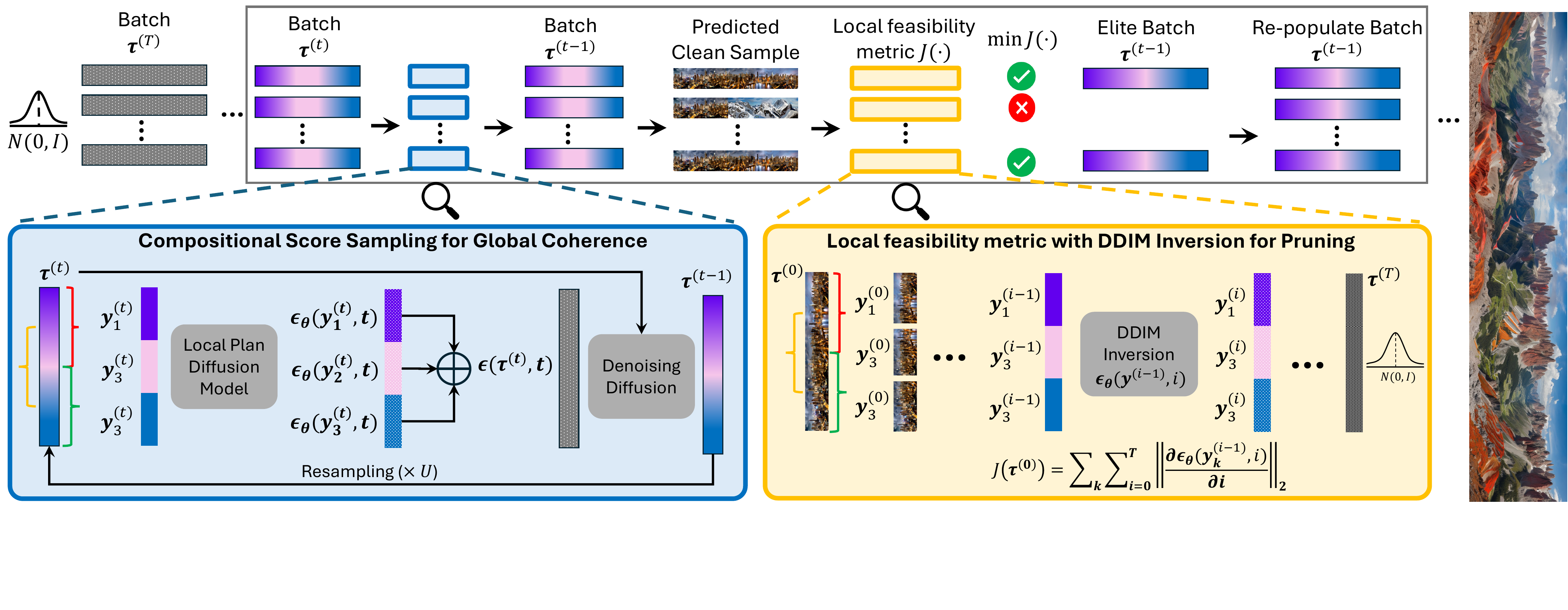}
    \caption{\textbf{Compositional diffusion with Guided Search.} At each denoising timestep, \ours{} iteratively denoises a batch of noisy candidate global plans by (i) \textbf{iterative resampling} to propagate information through averaged scores at overlaps (blue) and (ii) \textbf{pruning} candidates with local inconsistencies based on the predicted clean samples (yellow). This process ensures all local plans align and belong to high-likelihood regions of $p(y)$, producing globally coherent plans.}
    \label{fig:method}
    \vspace{-1em}
\end{figure}

\textbf{Our method: Compositional Diffusion with Guided Search (CDGS).} CDGS is a structured \emph{{inference-time} algorithm} designed to identify coherent sequences of local modes that form valid global plans. Specifically, CDGS employs a {population-based} search to explore and select promising mode sequences beyond na\"ive sampling. To facilitate the search, it: (i) incorporates iterative resampling into the compositional score calculation to enhance information exchange across distant segments, leading to potentially coherent global plan candidates, and (ii) prunes the incoherent candidates by evaluating the likelihood of their local segments with a ranking objective. Note that this is all within a standard denoising diffusion process, making CDGS a plug-and-play sampler applicable across domains, including robotics planning, panorama image generation, and long video generation. In the following sections, we detail each of these components and demonstrate how their integration enables efficient navigation of the complex multi-modal search space to produce coherent long-horizon plans.

\begin{figure}[t] 
\vspace{-1em}
\centering
\hbox{
\noindent
\footnotesize
\scalebox{0.87}
{
\begin{minipage}[t]{0.57\textwidth}

\begin{algorithm}[H]
\caption{\ours}
\label{algo:search_algo}
\begin{algorithmic}[1]
\Require Start $x_{s}$, Goal $x_{g}$, Planning horizon $H$
\Require Diffusion noise schedule, 
\Require Pretrained local plan score function $\epsilon_\theta(y^{(t)},t)$, 
\Require number of candidate plans $B$, number of elite plans $K$ at every step
\State Initialize $B$ global plan candidates: $\tau^{(T)}$
\State $\tau^{(T)} = (y^{(T)}_1 \circ \dots \circ y^{(T)}_M) \sim \mathcal{N}(0, \textbf{I})$
\For{$t = T, \dots, 1$}
    \State $\epsilon(\tau^{(t)},t)$ = ComposedScore($\tau^{(t)}, t, \epsilon_\theta, x_{s}, x_{g}$)
    \State $\hat{\tau}_0^{(t)} = (\tau^{(t)} - \sqrt{1-\alpha_t} \epsilon(\tau^{(t)},t))/\sqrt{\alpha_t}$
    \State Rank plans using $J(\hat{\tau}_0^{(t)})$ ~\autoref{eq:pruning_objective}
    \State Select best-$K$ global plans
    \State Repopulate candidates using filtered plans
    \State $\tau^{(t-1)} \sim p(\tau^{(t-1)}|\tau^{(t)}, \hat{\tau}_0^{(t)})$ ~\autoref{eq:tweedie}
\EndFor
\State return $\tau^{(0)}$
\end{algorithmic}
\end{algorithm}
\end{minipage}
}
\hfill
\scalebox{0.925
}{
\begin{minipage}[t]{0.52\textwidth}
\begin{algorithm}[H]
\caption{ComposedScore}
\label{algo:comp_algo}
\begin{algorithmic}[1]
\Require Noisy sample $\tau^{(t)}$, denoising timestep $t$, pretrained local plan score function $\epsilon_\theta$
\Require Start and goal: $x_{s}, x_{g}$
\Require Number of resampling steps $U$
\For{$u=1,\dots,U$}
    \State Calculate $\epsilon(\tau^{(t)},t)$ using~\autoref{eq:compositional_score}
    \If{$u < U$}
        \State Calculate $\tau^{(t-1)}$ using~\autoref{eq:tweedie}
        \State Add noise to $x_{s}$/$x_{g}$: 
        \State {\small$x^{(t-1)}_{s/g} \sim \mathcal{N}(\sqrt{\alpha_{t-1}}x_{s/g},  (1-\alpha_{t-1})I)$}
            \State Inpaint noisy start and goal in $\tau^{(t-1)}$
        \State Resampling: $\tau^{(t)}\sim p(\tau^{(t)}|\tau^{(t-1)})$ 
    \EndIf
\EndFor
\State return $\epsilon(\tau^{(t)},t)$
\end{algorithmic}
\end{algorithm}
\end{minipage}
}
\vspace{-1em}
}
\end{figure}

\subsection{Compositional Diffusion with Guided Search}


A key challenge with multi-modal distributions is that na\"ive compositional sampling can lead to incoherent global plans:
since each segment is independently sampled from $p(y)$, they may not align well at their overlaps and potentially lead to mode-averaging issues-where high-likelihood local plans do not combine to form a feasible global plan.

To address this, our approach leverages a guided search procedure that explores promising sequences of local modes while filtering out ones that are more likely to result in incoherent global plans.

\textbf{Method formulation.} At each diffusion timestep $t$, given a noisy global plan $\tau^{(t)}$, our goal is to sample from an improved next-step distribution over $\tau^{(t-1)}$, that is more likely to yield a coherent global plan. To achieve this, we define a modified sampling distribution:
\[\small
p_J\Big(\tau^{(t-1)}|\tau^{(t)}\Big) \propto p\Big(\tau^{(t-1)}|\tau^{(t)}\Big) \exp{\Big(- J\Big(\hat{\tau}_0^{(t-1)}}\Big)/\lambda_t\Big),
\]
where (i) $p(\tau^{(t-1)}|\tau^{(t)})$ is the original diffusion transition realized using the compositional score function $\epsilon(\tau^{(t)},t)$, (ii) $\hat{\tau}_0^{(t-1)}$ is the Tweedie-estimate of the clean global plan at timestep $t-1$, (iii) $J(\cdot)$ is a plan ranking metric we define below, and (iv) $\lambda_t$ controls the exploration-exploitation tradeoff. We approximate sampling from this distribution using a Monte Carlo search procedure resembling the cross-entropy
method: draw a batch of noisy global plans from $p(\tau^{(t-1)}|\tau^{(t)})$, rank them using $J$ and retain a subset of \textit{elite} global plans that minimizes the evaluation metric $J(\cdot)$ as illustrated in Algorithm~\ref{algo:search_algo}. The number of elites $K$ is a tunable parameter of our algorithm, enabling exploration of many possibilities in parallel when the planning problem is very large/difficult. Now, we just need to ensure that (i) the global plans are ranked appropriately and (ii) the candidate samples proposed by compositional sampling contain informative, globally coherent mode-sequences to pursue.

\textbf{Ranking global plans via local feasibility.} To guide the search effectively, we require a mechanism to evaluate the feasibility of candidate plans. Our key insight is that a global plan is feasible \emph{iff} all of its local transitions are feasible. Since the local model~$p(y)$ is trained to model feasible short-horizon behavior, high-likelihood local plans are strong indicators of local feasibility. Therefore, a globally feasible plan should consist of high-likelihood local-plan segments throughout. However, computing exact likelihoods in diffusion models is computationally expensive~\cite{song2020score}, often intractable. 

To address this, we leverage DDIM inversion~\cite{song2020ddim} to approximate the likelihoods of local {plan segments} $y$. Each local {segment} $y$ of a sampled global plan $\tau$ goes through forward diffusion \emph{using the learned score network}~($\epsilon_\theta$) such that:
\begin{equation}\small
\frac{y^{(t)}}{\sqrt{\alpha_t}} = \frac{y^{(t-1)}}{\sqrt{\alpha_{t-1}}} + \left(\sqrt{\frac{1 - \alpha_t}{\alpha_t}} - \sqrt{\frac{1 - \alpha_{t-1}}{\alpha_{t-1}}}\right) \epsilon_\theta(y^{(t-1)}, t)
\end{equation}
A high-likelihood sample follows a low-curvature path, whereas low-likelihood samples exhibit high curvature to bring noisy latents in-distribution when forward noised~\cite{heng2024out}~(refer~\autoref{app:sec:app-pruning}). Specifically, we define a smoothness measure based on the curvature of the diffusion trajectory during inversion:
\begin{equation}
\label{eq:pruning_objective} \small
    g\Big(y^{(0)}\Big) = \sum_{i=1}^{T} \Big\| \frac{\partial \epsilon_\theta(y^{(i-1)}, i)}{\partial i} \Big\|_2, \;\;\;\; J(\tau^{(0)}) = \prod_{m=1}^{M} \exp\Bigg(-g\Big(y_m^{(0)}\Big)\Bigg)
\end{equation}
where $g(y^{(0)})$ measures closeness of $y^{(0)}$ to the nearest mode of $p(y)$, intuitively. A higher value of $g(y^{(0)})$ corresponds to lower-likelihood local plans. We aggregate $g(y^{(0)})$ over all local plan segments $y^{(0)}_{1:M}$ in $\tau^{(0)}$ to define the global plan ranking metric $J(\tau^{(0)})$ to measure plan feasibility. Low-quality plans
have high $J$ values, making their denoising paths more likely to be pruned.

\subsection{Iterative Resampling}

To ensure the effectiveness of the guided search, it is not enough to rank global plans correctly—we must also promote 
globally coherent candidate plans.
However, standard compositional sampling fails to propagate long-horizon dependencies across overlapping local plans.
Consider the running example in \autoref{fig:multimodality}. After one denoising step, due to independent sampling of local plans, $y_1$ has no information about  $y_6$, and vice versa.

To address this, we apply iterative resampling ~\cite{lugmayr2022repaint}: repeatedly alternating between forward noising $\tau^{(t)}\sim p(\tau^{(t)}|\tau^{(t-1)})$ and denoising steps. This procedure enables the score network’s predictions for each segment to incorporate information from distant neighbors via overlapping variables, encouraging global consistency.
Mathematically, this process resembles belief propagation on a chain of factors where each local plan $y_m \in y_{1:M}$ in $\tau$ depends on its neighbors $y_{m-1}$ and $y_{m+1}$ through the respective overlaps~($y_m \cap y_{m-1}$ and $y_m \cap y_{m+1}$). During resampling, the belief of $y_m$ is updated as: $
p(y_m|y_{m-1}, y_{m+1}) \propto p(y_{m})p(y_m|y_m \cap y_{m-1}) p(y_m|y_m\cap y_{m+1})
$
Following Algorithm~\ref{algo:comp_algo}, after $U$ iterations, this iterative resampling ensures that information propagates across the entire long-horizon sequence, producing a more globally coherent plan.

\textbf{Summary of \ours{}.} We propose a 
guided-search algorithm by integrating a population-based pruning strategy within compositional sampling. Given a 
local plan score function, our approach samples potentially coherent global plan candidates 
and filters out plans with locally inconsistent segments. Repeating this throughout the denoising process improves the probability that the retained candidates satisfy local feasibility at every segment and are therefore globally feasible plans. Our algorithm benefits from adaptive compute at inference time, with the flexibility to scale the batch size $B$ and the number of resampling steps $U$ for problems with longer horizons and larger search spaces.
\section{Experimental Results: Robotic Planning}
\label{sec:results}

In this section, we evaluate the performance of \ours{} for long-horizon robotic planning.
For all the experiments, we represent inputs with a low-dimensional state-space of the system comprising the pose of the end-effector and the objects in the scene in the global frame of reference. For real-world evaluations, we obtain the pose of the objects through perception, more details in~\autoref{app:hardware}. 

\textbf{\ours{} can solve learning from play and stitching problems efficiently.}
We evaluate \ours{} for sequential-decision making tasks using the OGBench Maze and Scene task suite~\cite{park2024ogbench}, which includes PointMaze and AntMaze along with five tasks for Scene where a robot must manipulate objects (a drawer, sliding window, and cube) to reach a goal state. The primary challenge is learning from \emph{small maze trajcetories} or \emph{unstructured play data} during training, which does not directly solve the target tasks. The diversity of the unstructured plans makes the local distributions highly multi-modal. We hypothesize that \ours{} is an ideal method for this problem statement because it can compose short-horizon plans into meaningful long-horizon plans.

\begin{table}[h]
    \centering
    \resizebox{\linewidth}{!}{%
\footnotesize
\setlength{\tabcolsep}{5pt}
\begin{tabular}{ccc ccc ccc cccc}
\toprule
\textbf{Environment} & \textbf{Type} & \textbf{Size} & \textbf{GCBC} & \textbf{GCIVL} & \textbf{GCIQL} & \textbf{QRL} & \textbf{CRL} & \textbf{HIQL} & \textbf{GSC} & \textbf{CD} & \textbf{Ours {\tiny w/o PR}} & \textbf{Ours} \\
\midrule


\multirow[c]{3}{*}{\textbf{PointMaze}} 


 & \multirow[c]{3}{*}{\texttt{{Stitch}}} 

 & \texttt{Medium} & $23$ {\tiny $\pm 18$} & $70$ {\tiny $\pm 14$} & $21$ {\tiny $\pm 9$} & $0$ {\tiny $\pm 1$} & $80$ {\tiny $\pm 12$} & ${74}$ {\tiny $\pm 6$} 
 & \valstd{ \underline{100} } { 0 }
 & \valstd{ {100} } { 0 } & \valstd{ {100} } { 0 }  & \valstd{ \textbf{100} } { 0 }  \\

 &  & \texttt{Large} & $7$ {\tiny $\pm 5$} & $12$ {\tiny $\pm 6$} & $31$ {\tiny $\pm 2$} & $0$ {\tiny $\pm 0$} & $84$ {\tiny $\pm 15$} & ${13}$ {\tiny $\pm 6$} 
 & \valstd{ {100} } { 0 } 
 & \valstd{ {100} } { 0 } & \valstd{ \underline{100} } { 0 } & \valstd{ \textbf{100} } { 0 } \\

 &  & \texttt{Giant} & $0$ {\tiny $\pm 0$} & $0$ {\tiny $\pm 0$} & $0$ {\tiny $\pm 0$} & $0$ {\tiny $\pm 0$} & $50$ {\tiny $\pm 8$} & ${0}$ {\tiny $\pm 0$} 
 & \valstd{ {29} } { 2 }
& \valstd{ {68} } { 3 } & \valstd{ \underline{78} } { 2 } & \valstd{ \textbf{87} } { 3 }\\

\cmidrule{1-13}


\multirow[c]{3}{*}{\textbf{AntMaze}} 


 & \multirow[c]{3}{*}{\texttt{{Stitch}}} 

 & \texttt{Medium} & $45$ {\tiny $\pm 11$} & $44$ {\tiny $\pm 6$} & $29$ {\tiny $\pm 6$} & $59$ {\tiny $\pm 7$} & $53$ {\tiny $\pm 6$} & ${94}$ {\tiny $\pm 1$} 
 & \valstd{ \underline{97} } { 2 }
 & \valstd{ {96} } { 2 } & \valstd{ \underline{97} } { 2 }  & \valstd{ \textbf{97} } { 1 }  \\

 &  & \texttt{Large} & $3$ {\tiny $\pm 3$} & $18$ {\tiny $\pm 2$} & $7$ {\tiny $\pm 2$} & $18$ {\tiny $\pm 2$} & $11$ {\tiny $\pm 2$} & ${67}$ {\tiny $\pm 5$} 
 & \valstd{ {66} } { 2 } 
 & \valstd{ \underline{86} } { 2 } & \valstd{ \underline{86} } { 2 } & \valstd{ \textbf{88} } { 2 } \\

 &  & \texttt{Giant} & $0$ {\tiny $\pm 0$} & $0$ {\tiny $\pm 0$} & $0$ {\tiny $\pm 0$} & $0$ {\tiny $\pm 0$} & $0$ {\tiny $\pm 0$} & ${21}$ {\tiny $\pm 2$} 
 & \valstd{ {20} } { 1 }
& \valstd{ {65} } { 3 } & \valstd{ \underline{82} } { 1 } & \valstd{ \textbf{85} } { 3 }\\

\cmidrule{1-13}








\multirow[c]{3}{*}{\makecell{\textbf{Humanoid-}\\\textbf{Maze}}} 

 & \multirow[c]{3}{*}{\texttt{{Stitch}}} 
 & \texttt{Medium} & $29$ {\tiny $\pm 5$} & $12$ {\tiny $\pm 2$} & $12$ {\tiny $\pm 3$} & $18$ {\tiny $\pm 2$} & $71$ {\tiny $\pm 3$} & ${\textbf{96}}$ {\tiny $\pm 4$} 
 & \valstd{ {92} } { 1 }
 & \valstd{ {91} } { 1 }
  & \valstd{ {88} } { 3 } & \valstd{ \underline{93} } { 1 }
 \\
 &  & \texttt{Large} & $6$ {\tiny $\pm 3$} & $1$ {\tiny $\pm 1$} & $0$ {\tiny $\pm 0$} & $3$ {\tiny $\pm 1$} & $6$ {\tiny $\pm 1$} & ${31}$ {\tiny $\pm 3$}
 & \valstd{ {70} } { 3 }
 & \valstd{ \underline{72} } { 3 } 
 & \valstd{ {70} } { 2 } 
 & \valstd{ \textbf{74} } { 2 } 
 \\
 &  & \texttt{Giant} & $0$ {\tiny $\pm 0$} & $0$ {\tiny $\pm 0$} & $0$ {\tiny $\pm 0$} & $0$ {\tiny $\pm 0$} & $0$ {\tiny $\pm 0$} & ${12}$ {\tiny $\pm 2$} 
 & \valstd{ {5} } { 1 }
 & \valstd{ \textbf{64} } { 4 }
 & \valstd{ {47} } { 5 }
 & \valstd{ \underline{55} } { 3 }
 \\

\cmidrule{1-13}

\multirow[c]{1}{*}{\textbf{Scene}} 


 & \multirow[c]{1}{*}{\texttt{{Play}}}
 & - & $5$ {\tiny $\pm 1$} & $42$ {\tiny $\pm 4$} & \textbf{51} {\tiny $\pm 4$} & $5$ {\tiny $\pm 1$} & $19$ {\tiny $\pm 2$} & ${38}$ {\tiny $\pm 3$} 
 & \valstd{ {8} } { 2 }
 & \valstd{ {13} } { 1 } & \valstd{ \underline{36} } { 6 }  & \valstd{ \textbf{51} } { 2 }  \\

\bottomrule
\end{tabular}
}
\vspace{0.5em}
\caption{\textbf{OGbench~\cite{park2024ogbench}: learning from stitch and play datasets.} With much less training data requirements, \ours{} performs on-par with inverse-reinforcement learning baselines and better than generative baselines in a receding horizon control. \changes{For GSC, CD and \ours{}, we replan based on distance from goal for maze tasks (following CD~\cite{luo2025generative}) and sample the complete plan based on the oracle planning horizon for scene task. Success rate averaged over 100 trials and 3 seeds with randomly chosen task ids. Baseline performance is borrowed from original papers~\cite{park2024ogbench, luo2025generative}}}
    \label{tab:ogbench_table}
\vspace{-0.5em}

\end{table}

\begin{table}[t]

\centering
\resizebox{\linewidth}{!}{%
\footnotesize
\setlength{\tabcolsep}{5pt}
\begin{tabular}{ccc ccc ccc ccc cccc}
\toprule
\textbf{Environment} & \textbf{Size} & \textbf{GCBC} & \textbf{GCIVL} & \textbf{GCIQL} & \textbf{QRL} & \textbf{CRL} & \textbf{HIQL}  & \makecell{\textbf{GSC} \\ (4D)} &  \makecell{\textbf{CD} \\ (4D)}  
& \makecell{\textbf{GSC} \\ (17D)} & \makecell{\textbf{CD} \\ (17D)} & \makecell{\textbf{Ours {\tiny w/o PR}} \\ (17D)} & \makecell{\textbf{Ours} \\ (17D)}\\
\midrule
\multirow[c]{2}{*}{\makecell{\textbf{AntSoccer} }} 
 
 & \texttt{Arena} & ${34}$ {\tiny $\pm 4$} & $21$ {\tiny $\pm 3$} & $5$ {\tiny $\pm 2$} & $2$ {\tiny $\pm 1$} & $2$ {\tiny $\pm 1$} & $23$ {\tiny $\pm 2$} 
 & \valstd{ {41} } { 4 }
 & \valstd{ {55} } { 6 } 
 & \valstd{ {65} } { 3 }
 & \valstd{ \textbf{69} } { 3 }
  & \valstd{ \underline{67} } { 3 }
  & \valstd{ \textbf{69} } { 1 } \\
 & \texttt{Medium} & $2$ {\tiny $\pm 1$} & $1$ {\tiny $\pm 0$} & $0$ {\tiny $\pm 0$} & $0$ {\tiny $\pm 0$} & $0$ {\tiny $\pm 0$} & ${8}$ {\tiny $\pm 2$} 
 & \valstd{ {5} } { 2 } 
 & \valstd{ {13} } { 1 } 
 & \valstd{ {12} } { 2 } 
 & \valstd{ \underline{17} } { 3 }
& \valstd{ {16} } { 3 }
 & \valstd{ \textbf{18} } { 2 } \\
 
 
\bottomrule
\end{tabular}
}
\vspace{0.5em}
\caption{\textbf{Stitching composite task AntSoccer in OGBench~\cite{park2024ogbench}.} We evaluate \ours{} on high-dimensional (17D) state space to stitch ball reaching and ball carrying behaviors for two AntSoccer environments. Success rate averaged over 100 trials and 3 seeds with randomly chosen task ids. Baseline performance is borrowed from original papers~\cite{park2024ogbench, luo2025generative}}
\label{table:ant_soccer}
\vspace{-1em}
\end{table}

\begin{table}[b] \small
    \centering
    \spacing{-1em}
    \resizebox{\linewidth}{!}{%
    \begin{tabular}{lccccccc}
       \multirow{2}{*}{} & \multirow{2}{*}{\makecell{\textbf{Remark} \\\textbf{(Task information)}}} & \multicolumn{2}{c}{\textbf{Hook Reach}} & \multicolumn{2}{c}{\textbf{Rearrangement Push}} & \multicolumn{2}{c}{\textbf{Rearrangement Memory}}\\
       \cline{3-8}
       & & \textbf{Task 1} & \textbf{Task 2} & \textbf{Task 1} & \textbf{Task 2} & \textbf{Task 1} & \textbf{Task 2} \\
       \hline         
       Task Length & & 4 & 5 & 4 & 7 & 4 & 7\\
       \hline
        Random CEM & \multirow{2}{*}{PDDL + BFS} & 0.14 & 0.10 & 0.08 & 0.00 & 0.02 & 0.02 \\
        STAP CEM & & \textbf{0.66} & \textbf{0.70} & 0.76 & \textbf{0.70} & 0.00 & 0.00 \\
        \hline 
       \hline
       LLM-T2M, $n=11$ & \multirow{2}{*}{\makecell{LM Prior +\\ Prompting}} & 0.0 & 0.48 & 0.72 & 0.06 & 0.0 & 0.0 \\
       VLM-T2M, $n=11$ & & 0.0 & 0.42 & 0.62 & 0.02 & 0.0 & 0.0 \\
        \hline 
       \hline
       \changes{GSC (Original)} & \changes{\multirow{1}{*}{\makecell{Oracle task plan}}} & \changes{0.78} & \changes{0.80} & \changes{0.88} & \changes{0.64} & \changes{0.82} & \changes{0.48} \\
       \hline
       \hline
       GSC (no task plan) & \multirow{3}{*}{\makecell{No PDDL \\ skill-level data only}} & 0.18 & 0.04 & 0.00 & 0.00 & 0.07 & 0.00 \\
       \ours {} (w/o PR) & & 0.24 & 0.12 & 0.12 & 0.00 & 0.11 & 0.00 \\
       \ours {} (ours) & & \textbf{0.64} & 0.58 & \textbf{0.84} & 0.48 & \textbf{0.42} & \textbf{0.18}\\
        \hline 
    \end{tabular}
    }
    \caption{\textbf{Evaluation on TAMP task-suite.} We compare \ours{} with relevant search-based~(PDDL Domain) and prompting based~(LLM/VLM) baselines.
    \ours{} performs on-par or slightly trails privileged methods on Hook Reach and Rearrangement Push, but substantially outperforms them on Rearrangement Memory. (success rate over $50$ trials)
    }
    \label{tab:results_baseline}
\end{table}

\ours{} uses a Diffuser~\cite{janner2022diffuser} to learn the distribution of local plans (up to 4~secs of trajectory at 20~Hz) represented as a sequence of states and actions $y = \{s_1, a_1, ..., s_h, a_h\}$ and then composes them at inference for a given goal state to sample up to 10~secs of motion plans $\tau=\{s_i,a_i\}_{i=1}^{H}$~($h<H)$. \changes{We compare the performance of \ours{} with inverse reinforcement learning baselines from OGBench, including GCBC~\cite{lynch2020learning,ghosh2019learning}, GCIVL, GCIQL~\cite{kostrikov2021offline}, and HIQL~\cite{park2024hiql}, with results presented in~\autoref{tab:ogbench_table}. In addition we also include compositional generative baselines like GSC~\cite{mishra2023generative} and CompDiffuser~\cite{luo2025generative}.} It should be noted that CDGS with resampling and pruning can scale the performance of na\"ive compositional sampling~(GSC), in a training-free manner, to an extent that beats baselines like CompDiffuser~\cite{luo2025generative} that use overlap information while training and learn an overlap conditioned score function. Finally, we also validate \ours{} on composite ball reaching and ball carrying trajectory stitching of AntSoccer in OGBench and show the results in~\autoref{table:ant_soccer}.

\begin{figure}[t]
    \centering
    \includegraphics[width=\linewidth, trim={0cm 0cm 0cm 0.5cm}]{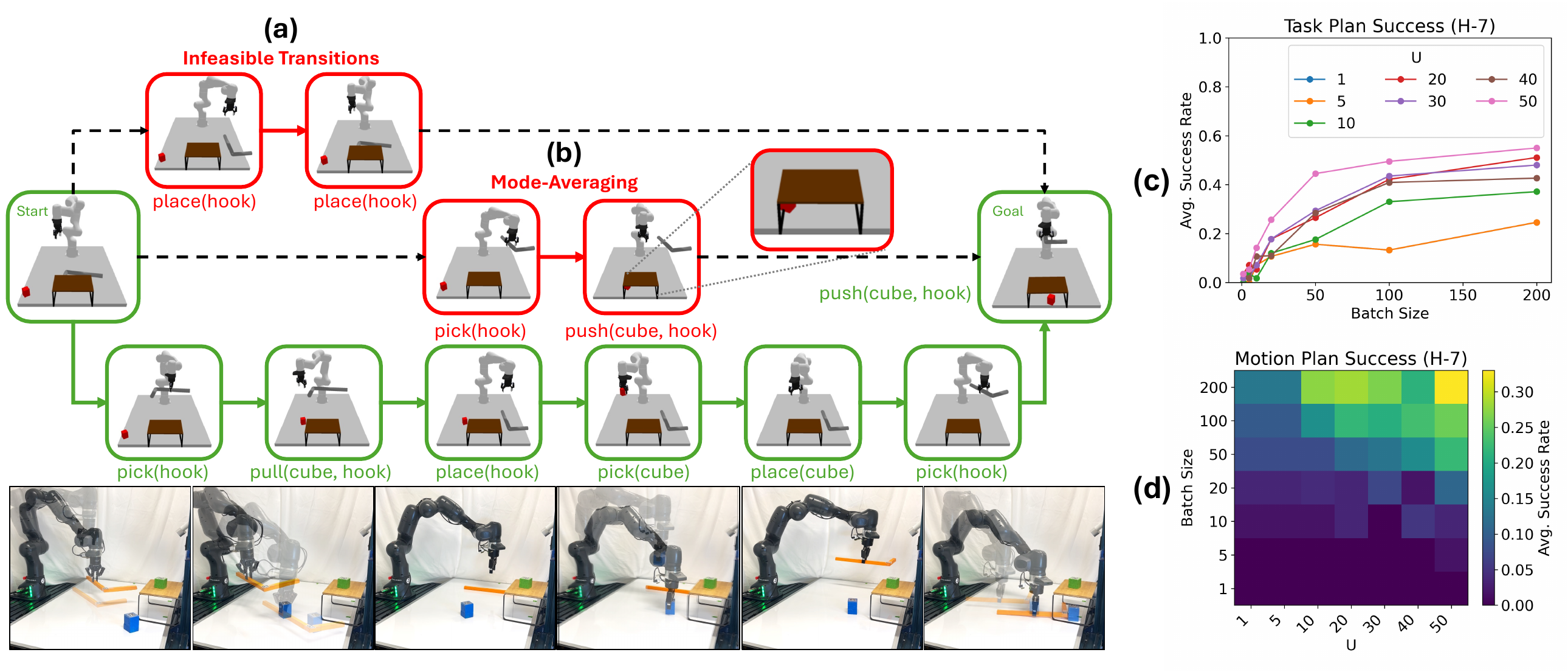}
    \caption{\textbf{Left: Visualizing plan pruning.} When compositional sampling chooses an infeasible mode sequence, the resulting plan can hallucinate out-of-distribution transitions due to \textbf{mode-averaging} as explained in~\autoref{sec:method}. For instance, \textbf{(a) Infeasible transitions:} \texttt{inhand(hook)} precondition is never met for \texttt{place(hook)}, and \textbf{(b) State hallucination:} \texttt{cube} moves \texttt{under(rack)} as a result of averaging toward the goal state, despite being geometrically infeasible for \texttt{push(cube, hook)}. Our pruning objective~(\autoref{eq:pruning_objective}) ensures only feasible plans during denoising, where all transitions are in-distribution with our short-horizon transition diffusion model.
    \textbf{Right: Scaling analysis.} 
    (H-7) denotes performance averaged over tasks of horizon 7.
    \textbf{(c) Task planning} success improves with batch size, with larger gains from more resampling steps.
    \textbf{(d) Motion planning} success improves with resampling steps, but only when batch size is large enough
    }
    \label{fig:task_success_vs_batch_size}
    \spacing{-1em}
\end{figure}


    

\textbf{\ours{} can solve hybrid-planning problems.}
Task and Motion Planning (TAMP) decomposes robotic planning into a symbolic search for a sequence of discrete high-level skills (e.g., \texttt{pick}, \texttt{place}, \texttt{pull}) followed by low-level motion planning for each skill~\cite{garrett2021integrated}. Specifically, we formulate a task-and-motion-plan as $\tau = \{y_1, \ldots, y_m\}$ where $y_i=\{s_i,\pi_i,a_i,s_{i+1}\}$ where a discrete skill $\pi_i$ with motion parameters $a_i$ is executed on state $s_i$ to get to state $s_{i+1}$. This entails solving a hybrid-planning problem where the chosen discrete skill modes and continuous action modes must simultaneously satisfy symbolic and geometric constraints.
We systematically evaluate our method on three suites of TAMP tasks, which are described in detail in ~\autoref{app:tasks}. 

\changes{We compare our method with learning-based TAMP and other compositional methods. Specifically, we consider the following categories: (1) privileged with Planning Domain Definition Language~(PDDL): \textbf{Random CEM} and \textbf{STAP CEM}~\cite{agia2023stap} search for symbolic plans in a manually and systematically constructed PDDL domain and apply Cross Entropy Method~(CEM) optimization over
potential motion plans 
(2) task information provided via prompting: \textbf{LLM-T2M}~\cite{lin2023text2motion} prompts an LLM (\texttt{GPT-4.1}) and VLM (\textbf{VLM-T2M}) with descriptions of the scene along with $n = 11$ in-context examples~(w/o and w/ scene images respectively) to generate a feasible task plan that is checked by a geometric motion planner (STAP CEM in this case). 
(3) compositional diffusion: \textbf{GSC (no task plan)}~\cite{mishra2023generative} performs compositional diffusion (equivalent to \ours{} w/o RP and PR). Notably, GSC and \ours{} are the \textit{only} methods that do not rely on explicit symbolic search or LLM/VLM supervision for the task plan. The results of our evaluation are in \autoref{tab:results_baseline}. Note that while \textbf{GSC (Original)}~\cite{mishra2023generative} leverages skill-level expert diffusion models and oracle task plan, in our case it represents na\"ive compositional sampling with a unified model (w/o oracle task plan).}


\textbf{\ours{}'s performance scales with compute.}
We hypothesize that \ours{} has adaptive inference-time compute, meaning that it benefits from more compute on harder problems. We validate this hypothesis on our most challenging TAMP tasks with a planning horizon of 7. 
We find that increasing batch size~($B$) and number of resampling steps ~($U$) increases the task planning success ~\autoref{fig:task_success_vs_batch_size}(c) and motion planning success  ~\autoref{fig:task_success_vs_batch_size}(d) of \ours. 
Interestingly, we find that neither increasing $B$ nor $U$ on their own is sufficient for overall motion planning success.
Thus, both resampling and pruning are essential for long-horizon tasks, as evidenced by the significant improvement of \ours{}~( \autoref{tab:results_baseline}).

\section{\ours{} for Long content generation}
\label{sec:results_imgvideo}

We formulate \ours{} with specific design choices that enable (i) efficient message passing for global consistency and (ii) pruning denoised paths that lead to incoherent sequences. While these mechanisms are essential for long-horizon planning, we investigate their broader applicability, particularly in long-content generation tasks such as text-to-image~(T2I) and text-to-video~(T2V), which require spatial and temporal coherence over extended horizons. Our framework demonstrates effective improvement in long-horizon content generation.

\begin{figure}[t]
    \centering
    \includegraphics[width=\linewidth, trim={0cm 0cm 0cm 0cm}]{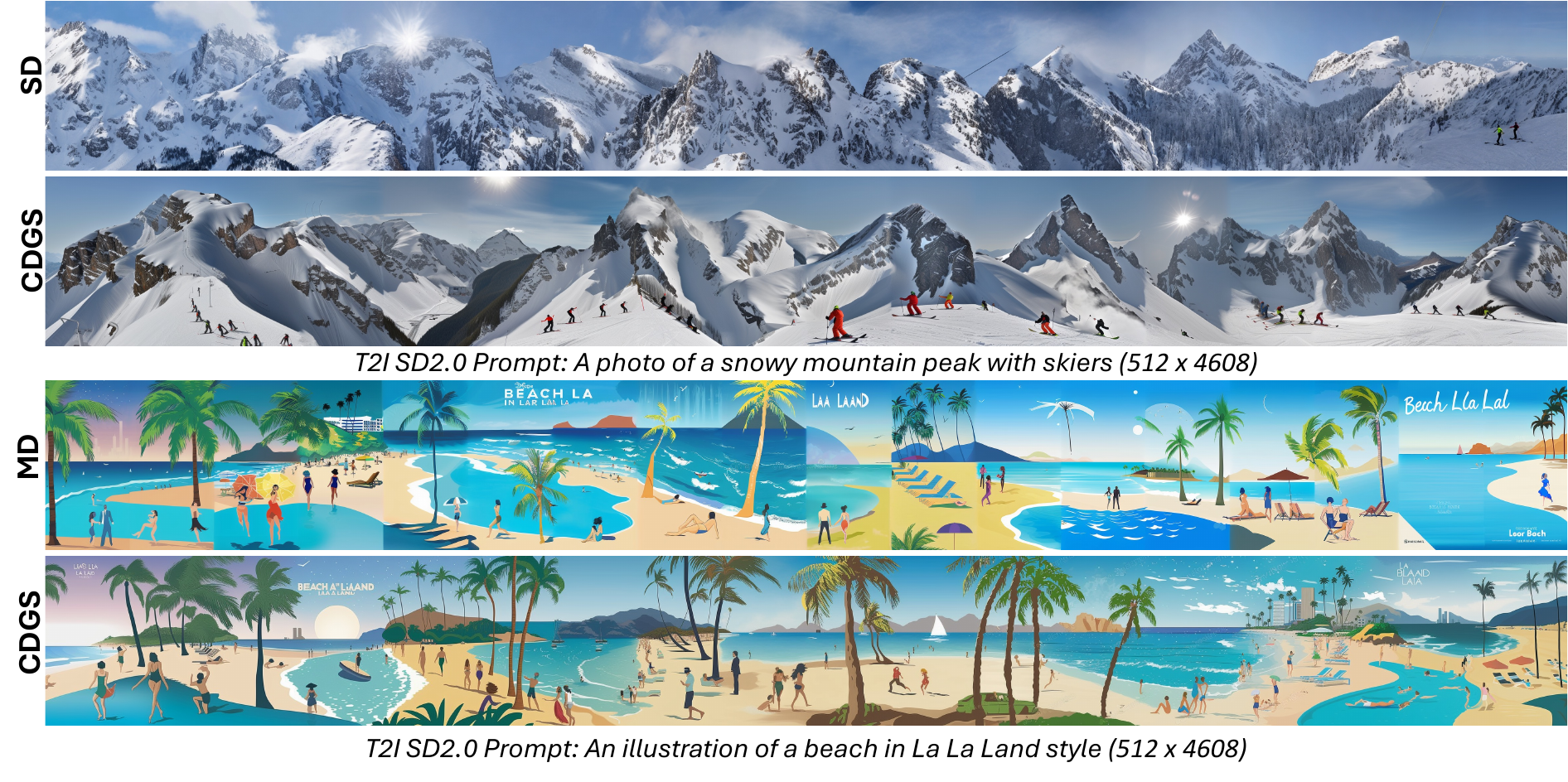}
    \caption{\textbf{Panorama image generation.} The above figure shows the qualitative comparison of \ours{} with MD~\cite{bar2023multidiffusion} and SD~\cite{lee2023syncdiffusion}. We show qualitative intuition behind global coherence and local feasibility: while SD generates smooth panoramas, they fail to satisfy the global context (\textit{mountain peak with skiers}), on the other hand, MD follows the global context (\textit{beach in La La Land style}) but fails to exhibit local consistency. \ours{} excels at both.
    }
    \label{fig:imggen_exp}
\end{figure}

\textbf{\ours{} enables coherent panoramic image generation via stitching.} We evaluate \ours{} on panoramic synthesis by composing multiple image patches. A panorama $\tau$ is represented as a sequence of small images $y$, each split into three overlapping patches $y = (x_1, x_2, x_3)$. Using Stable Diffusion-2.0~\cite{stable_diffusion2}, we generate up to $512\times 4608$ panoramas by stitching $512\times 512$ images. We compare against (i) Multi-Diffusion~(MD)~\cite{bar2023multidiffusion}, which averages scores across overlaps (image-domain analogue of GSC~\cite{mishra2023generative}), and (ii) Sync-Diffusion~(SD)~\cite{lee2023syncdiffusion}, which enforces LPIPS-based perceptual guidance~\cite{zhang2018unreasonable}. As shown in~\autoref{tab:panorama_table}, \ours{} matches SD without explicit perceptual loss, indicating effective message passing for global style and perceptual transfer while maintaining prompt alignment (CLIP~\cite{radford2021learning}). Qualitative samples are shown in~\autoref{fig:imggen_exp}, with more details in~\autoref{app:sec:imggen}.

\begin{table}[h]
    \centering
    \resizebox{0.9\linewidth}{!}{%
    \begin{tabular}{lcccc}
 Metric & \textbf{GSC/Multi-Diffusion} & \textbf{ Sync-Diffusion} & \textbf{\ours{} w/o PR} & \textbf{\ours{}} \\
\hline
{Intra-LPIPS} $\downarrow$
 & $0.72$ {\tiny $\pm 0.08$} & $\mathbf{0.58}$ {\tiny $\pm \mathbf{0.06}$} & $0.61$ {\tiny $\pm 0.08$} & $\mathbf{0.59}$ {\tiny $\pm \mathbf{0.04}$} \\
{Intra-Style-L{\tiny$(\times 10^{-2})$}} $\downarrow$
 & $2.96$ {\tiny $\pm 0.24$} & $\mathbf{1.39}$ {\tiny $\pm \mathbf{0.12}$} & $1.97$ {\tiny $\pm 0.08$} & $\mathbf{1.38}$ {\tiny $\pm \mathbf{0.03}$} \\
{Mean-CLIP-S} $\uparrow$
 & ${31.77}$ {\tiny $\pm 2.14$} & ${31.77}$ {\tiny $\pm 2.14$} & ${31.71}$ {\tiny $\pm 2.34$} & $\mathbf{32.51}$ {\tiny $\pm \mathbf{2.66}$} \\


 \hline
\end{tabular}
}
    \caption{\textbf{Quantitative comparison of panorama generation.} We generate 1000 panoramas of dimensions $512\times4608$ using 14 prompts and compare different methods based on their perceptual similarity~(LPIPS~\cite{zhang2018unreasonable}), style similarity~(Style-loss~\cite{style_loss}), and prompt alignment~(CLIP score~\cite{radford2021learning}).}
    \label{tab:panorama_table}
    \vspace{-1em}
\end{table}


\begin{figure}[t]
    \centering
    \includegraphics[width=\linewidth, trim={0cm 1cm 0cm 0cm}]{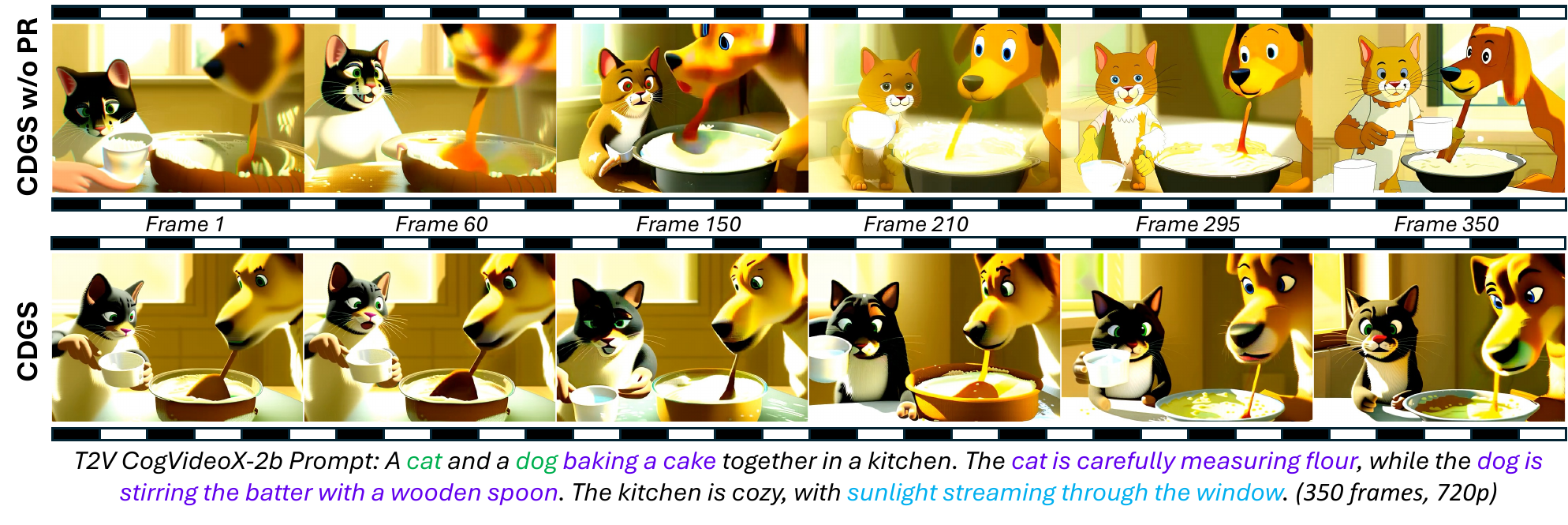}
    \caption{\textbf{Long video generation.} \ours{} w/ PR~(below) maintains subject-consistency while \ours{} w/o PR~(top) exhibits mode-averaging, resulting in significant changes to the subjects' appearances.
    }
    \label{fig:vidgen_exp}
\end{figure}

\textbf{\ours{} can sample temporally-consistent longer videos.} We follow a setup similar to panorama generation, composing shorter clips along the temporal axis for long-video generation. When short sequences of frames are stitched to make a long video, a key challenge is maintaining subject consistency and minimizing temporal artifacts. We use CogVideoX-2B~\cite{yang2024cogvideox} as the base model, capable of generating $\sim 50$-frame videos, and extend it to up to 350 frames at 720p resolution. We use six prompts to generate videos with na\"ive composition~(GSC/Gen-L-Video~\cite{wang2023gen} equivalent), compositional diffusion with resampling, and \ours{}. The results are evaluated with VBench~\cite{huang2024vbench} for temporal consistency, subject fidelity, visual quality, and alignment with the prompt~(refer~\autoref{tab:videogen_table}). Qualitative analysis in~\autoref{fig:vidgen_exp} clearly shows the multimodal problem where multiple local plans allow satisfying the global context, but with \ours{}'s effect local-to-global message passing, we see an improvement in subject consistency and temporal smoothness. This comes at a minor aesthetic degradation—a tradeoff commonly observed in long-video generation models. 



\begin{table}[h]
    \centering
    \resizebox{\linewidth}{!}{%
    \begin{tabular}{ccccc}
\textbf{Method} & \textbf{Subject-consistency} $\uparrow$ & \textbf{Temporal-flickering} $\uparrow$ & \textbf{Aesthetic-quality} $\uparrow$ & \textbf{Prompt-alignment} $\uparrow$ \\
\hline
CogVideoX-2B (50 frames)
 & $\mathbf{95.91}$ & $\mathbf{97.35}$ & $\mathbf{63.10}$ & $\mathbf{25.51}$\\
 CogVideoX-2B (350 frames) & $90.24$ & $98.44$ & $49.44$ & $21.78$\\
 \hline
GSC ($\equiv$ Gen-L-Video)
 & $89.51$ & $96.89$ & $\mathbf{60.12}$& $25.13$ \\
 Ours w/o PR
 & $91.06$ & $97.08$ & $59.40$ & $25.42$\\
  Ours
 & $\mathbf{91.67}$ & $\mathbf{97.16}$ & $58.90$ & $\mathbf{26.13}$\\


 \hline
\end{tabular}
}
    \caption{\textbf{Quantitative comparison of long-video generation.} We evaluate the performance of \ours{} based on selected metrics from VBench that measure subject consistency, aesthetics, prompt alignment and temporal artifacts. We use 6 prompts~(refer \autoref{app:sec:vidgen}) and generate videos with 350 frames at 720p resolution. \ours{} achieves competitive video quality but for significantly (7x) extended horizons.}
    \label{tab:videogen_table}
    \vspace{-1em}
\end{table}



\section{Related Work}
\label{sec:literature}
\textbf{Long-horizon content generation}
There are many approaches to generating long-horizon content like panoramas and long videos \cite{kalischek2025cubediff, lu2024freelong, chen2023seine, henschel2025streamingt2v}. Some assume access to long-horizon training data for end-to-end training \cite{ge2022long, chen2024forcing, xie2025progressive, yan2025long}, while others with weaker assumptions about training data will compose the outputs of short-horizon models through outpainting \cite{wu2022nuwa, kim2024fifo} or stitching \changes{\cite{zhang2023diffcollage, kim2025syncos, li2024diffstitch, lee2025state, pan2024t, chen2025extendable, luo2025generative}}. Our method belongs to the latter, enabling generalization to longer horizons than seen during training.

\textbf{Generative planning.}
Generative models such as diffusion models\cite{sohl-dickstein_deep_2015, ho2020denoising} are widely used for planning~\cite{janner22diffuser, ajay2023decisiondiffuser, chi2023diffusionpolicy, li2023hierarchicaldiffusion, luo2025generative}, though they struggle with task lengths beyond their training data. Recent works including Diffusion-CCSP \cite{yang2023compositional}, GSC \cite{mishra2023gsc}, and GFC \cite{mishra2024gfc} have explored compositional sampling~\cite{liu2022compositional, du2023reduce,zhang2023diffcollage} but 
they sidestep the mode-averaging problem via additional mode supervision in the form of task skeletons or constraint graphs.
In contrast, our approach
directly addresses the mode-averaging problem to
generate goal-directed long-horizon plans 
from short-horizon models.

\textbf{Inference-time compute.}
Scaling inference-time computation is a powerful strategy for improving the performance of generative models \cite{wei2022chain, muennighoff2025s1}. 
For diffusion models \cite{song2020ddim, karras2022edm}, recent work has shown the efficacy of scaling inference-time compute through verifier-guided search during the denoising process \cite{ma2025inference,singhal2025general,yoon2025monte, zhang2025t, zhang2025inference}. Our algorithm differs in that it addresses the unique limitation of mode-averaging when sampling from a compositional chain of distributions.
\section{Conclusion}
\label{sec:conclusion}


We introduce \ours{}, a framework integrating compositional diffusion with guided search to generate long-horizon sequences with short-horizon models. By embedding search within the denoising process, \ours{} can handle composing highly multimodal distributions and sample solutions that are both globally coherent and locally feasible. Qualitative and quantitative results suggest that \ours{} is a general pathway for extending the reach of generative models beyond their training horizons across robotic planning, panoramic images, and video generation.

\section{Limitations}
\label{sec:limitations}


While \ours{} demonstrates strong performance in long-horizon goal-directed planning, it relies on a few simplifying assumptions that also suggest directions for future work. We assume the ability to specify a goal state, which simplifies planning but can be naturally extended to goal-generation or classifier-guided goal-conditioning methods~\cite{florensa2018goalgeneration}. Similarly, we generate plans for a fixed horizon, yet the framework can handle arbitrary horizons given the same start and goal, enabling selection among multiple candidate plan lengths. Finally, long-horizon dependencies are communicated through score averaging and resampling between adjacent skills; more sophisticated message-passing or attention-based mechanisms could improve efficiency and coherence across entire plans. These assumptions keep the problem tractable while providing a flexible foundation for extending \ours{} to more general and complex planning scenarios.

\clearpage

\section{Reproducibility statement}

We are committed to ensuring that all the results presented in this paper are reproducible. To this end, we have provided pseudocodes in the paper and released the official code base through our project website: \url{https://cdgsearch.github.io/}. We have also provided the hyperparameters table for motion planning~(refer~\autoref{app:hparams}), for image generation~(refer~\autoref{app:sec:image-code}) and video generation~(refer~\autoref{app:sec:video-code}). Apart from this our content-generation experiments use open-source models like Stable-Diffusion-2~(refer \url{https://huggingface.co/stabilityai/stable-diffusion-2}) and CogVideoX-2B~(refer \url{https://huggingface.co/zai-org/CogVideoX-2b}). For all other robotics setup, we provide more information through appendix and our project website.

\begin{rebuttal}
\section{LLM Usage}

LLMs were not used in any manner for conceptualization of the idea, key contributions of the proposed work and finding relevant prior woks.
\end{rebuttal}


\section*{Acknowledgments}
This work is partially supported by NSF-2442393, NSF-2409016, NSF-1942523, and Samsung Research America.

\bibliography{iclr2026_conference}
\bibliographystyle{iclr2026_conference}

\newpage

\appendix
{
\begingroup
\hypersetup{linkcolor=black}
\tableofcontents
\endgroup
}
\newpage

\section{Additional Panorama Generation Results}
\label{app:sec:imggen}

\begin{figure}[h]
\centering
    \includegraphics[width=\linewidth]{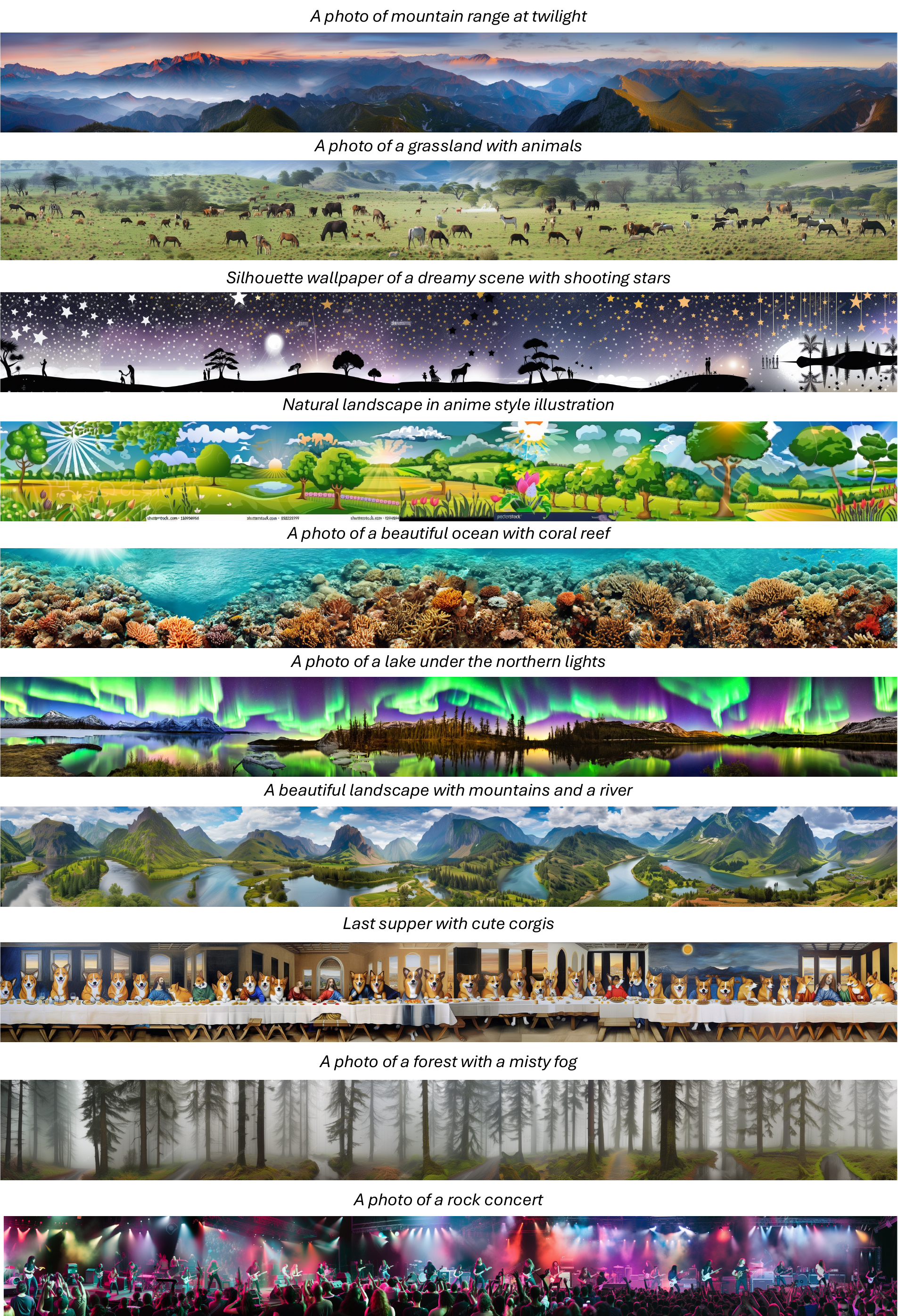}
\end{figure}

\newpage

\section{Prompts for Video Generation}
\label{app:sec:vidgen}

We used the following standard prompts for generating the videos:
\begin{enumerate}
    \item The camera follows behind a white vintage SUV with a black roof rack as it speeds up a steep dirt road surrounded by pine trees on a steep mountain slope, dust kicks up from it's tires, the sunlight shines on the SUV as it speeds along the dirt road, casting a warm glow over the scene. The dirt road curves gently into the distance, with no other cars or vehicles in sight. The trees on either side of the road are redwoods, with patches of greenery scattered throughout. The car is seen from the rear following the curve with ease, making it seem as if it is on a rugged drive through the rugged terrain. The dirt road itself is surrounded by steep hills and mountains, with a clear blue sky above with wispy clouds. realism, lifelike.
    
    \item A cute happy panda, dressed in a small, red jacket and a tiny hat, sits on a wooden stool in a serene bamboo forest. The panda's fluffy paws strum a miniature acoustic guitar, producing soft, melodic tunes, move hands, singings. Nearby, a few other pandas gather, watching curiously and some clapping in rhythm. Sunlight filters through the tall bamboo, casting a gentle glow on the scene. The panda's face is expressive, showing concentration and joy as it plays. The background includes a small, flowing stream and vibrant green foliage, enhancing the peaceful and magical atmosphere of this unique musical performance. realism, lifelike.
    
    \item A group of colorful hot air balloons take off at dawn in Cappadocia, Turkey. Dozens of balloons in various bright colors and patterns slowly rise into the pink and orange sky. Below them, the unique landscape of Cappadocia unfolds, with its distinctive 'fairy chimneys' - tall, cone-shaped rock formations scattered across the valley. The rising sun casts long shadows across the terrain, highlighting the otherworldly topography. realism, lifelike.

    \item A detailed wooden toy ship with intricately carved masts and sails is seen gliding smoothly over a plush, blue carpet that mimics the waves of the sea. The ship's hull is painted a rich brown, with tiny windows. The carpet, soft and textured, provides a perfect backdrop, resembling an oceanic expanse. Surrounding the ship are various other toys and children's items, hinting at a playful environment. The scene captures the innocence and imagination of childhood, with the toy ship's journey symbolizing endless adventures in a whimsical, indoor setting. realism, lifelike. 
    
    \item A young woman with beautiful and clear eyes and blonde hair standing and white dress in a forest wearing a crown. She seems to be lost in thought, and the camera focuses on her face. The video is of high quality, and the view is very clear. High quality, masterpiece, best quality, highres, ultra-detailed, fantastic. realism, lifelike.
    
    \item A woman walks away from a white Jeep parked on a city street at night, then ascends a staircase and knocks on a door. The woman, wearing a dark jacket and jeans, walks away from the Jeep parked on the left side of the street, her back to the camera; she walks at a steady pace, her arms swinging slightly by her sides; the street is dimly lit, with streetlights casting pools of light on the wet pavement; a man in a dark jacket and jeans walks past the Jeep in the opposite direction; the camera follows the woman from behind as she walks up a set of stairs towards a building with a green door; she reaches the top of the stairs and turns left, continuing to walk towards the building; she reaches the door and knocks on it with her right hand; the camera remains stationary, focused on the doorway; the scene is captured in real-life footage.
    
    \item At sunset, a modified Ford F-150 Raptor roared past on the off-road track. The raised suspension allowed the huge explosion-proof tires to flip freely on the mud, and the mud splashed on the roll cage.
    
    \item A cat and a dog baking a cake together in a kitchen. The cat is carefully measuring flour, while the dog is stirring the batter with a wooden spoon. The kitchen is cozy, with sunlight streaming through the window.

\end{enumerate}

\section{Compositional Score Computation: \ours{}'s relation to existing literature}
\label{app:score_composition}


Composing the distributions defined by multiple diffusion models is well-explored in literature \cite{du2023reduce, yang2023compositional}. Specifically we want to sample from the distribution of long-horizon sequences $\tau =(x_1, x_2, ..., x_N)$ by composing distributions of short-horizon sequences. There have been two main ways of composing short-horizon diffusion models in a chain:

\begin{enumerate}
    \item Score-Averaging: approaches like GSC~\cite{mishra2023gsc} and \ours{} partition $\tau$ into overlapping segments where the score for regions of overlap can be obtained by score-averaging:
    \[
    p(\tau) \propto \frac{p(x_1, x_2, x_3)p(x_3, x_4, x_5) \ldots}{p(x_3) \ldots}
    \]
    \item Conditioning: CompDiffuser~\cite{luo2025generative} partitions $\tau$ into non-overlapping segments that are conditioned on adjacent segments
    \[
    p(\tau) \propto p(x_1|x_2)p(x_N|x_{N-1})\prod_{i=2}^{N-1} p(x_i | x_{i-1}, x_{i+1})
    \]
\end{enumerate}

Since CompDiffuser~\cite{luo2025generative} requires training a model with conditions, we follow the more plug-n-play format of GSC~\cite{mishra2023generative}.
For TAMP, the key difference between \ours{} and GSC is that individual skill-level transitions for GSC are already conditioned on the task plan. This means that \ours{} samples from the unified model $p(s_{i-1}, a_i, s_i)$ where for GSC individual segments are sampled from $p(s_{i-1}, a_i, s_i| \pi_i)$ since the oracle skill-sequence (task plan) $\pi_{1:H}$ is already provided. This greatly simplifies compositional sampling as the models in GSC only conduct motion planning, thus reducing multi-modality and mode-averaging issues significantly, whereas the models in \ours{} conduct full task and motion planning.

\section{Pruning objective via DDIM Inversion: \ours{}'s relation to existing literature} \label{app:sec:app-pruning}

DDIM Inversion is simply running the DDIM~\cite{song2020ddim} denoising process backward i.e., forward noising in a deterministic way, to extract the denoising path from clean samples. Since we sample plans from a composed distribution, transition segments of a good plan should follow high-likelihood regions of the unified skill-transition distribution. A DDIM sampling based denoising looks like:
\[
x^{(t-1)} = \sqrt{\alpha_{t-1}}\left(\frac{x^{(t)} - \sqrt{1 - \alpha_{t}}\epsilon_\theta(x_t, t)}{\sqrt{\alpha_t}}\right) + \sqrt{1 - \alpha_{t-1}}\epsilon_\theta(x^{(t)}, t)
\]
We follow~\cite{heng2024out} to formulate this metric by first forward-noising each segment of the sampled plan from the task-level distribution according to:
\[
\frac{x^{(t)}}{\sqrt{\alpha_t}} = \frac{x^{(t-1)}}{\sqrt{\alpha_{t-1}}} + \left(\sqrt{\frac{1 - \alpha_t}{\alpha_t}} - \sqrt{\frac{1 - \alpha_{t-1}}{\alpha_{t-1}}}\right) \epsilon_\theta(x^{(t-1)}, t)
\]
With $\delta_t = \sqrt{\frac{1 - \alpha_t}{\alpha_t}}$ and $y^{(t)} = x^{(t)} \sqrt{1 + \delta_t^2}$, we can convert the above into:
\[
dy_t = \epsilon_\theta(x^{(t-1)}, t) d\delta_t
\]

Lets consider two forward-noising paths from two samples: one from high-likelihood region and one from a low-likelihood region. For both the samples, the rate of change of the integration path and its curvature directly indicate the likelihood of the clean sample. A high-likelihood sample will follow a smoother path with less curvatures while a low -likelihood sample will follow a high-curvature path to bring the noisy samples to high-likelihood regions of the noisy distribution. Hence, we consider Taylor expansion to analyze the higher order terms:
\begin{align*}
    y^{(t+1)} &= y^{(t)} + (\delta_{t+1} - \delta_t) \frac{dy^{(t)}}{d\delta_t}|_{(y^{(t)}, t)} + (\delta_{t+1} - \delta_t)^2 \frac{d^2y^{(t)}}{d\delta_t^2}|_{(y^{(t)}, t)} + \dots \\
    &= y^{(t)} + (\delta_{t+1} - \delta_t) \epsilon_\theta(x^{(t-1)}, t) + (\delta_{t+1} - \delta_t)^2 \frac{d\epsilon_\theta(x^{(t-1)}, t)}{d\delta_t}|_{(y^{(t)}, t)} + \dots
\end{align*}

where the second derivative term can be further decomposed into
\[
\frac{d\epsilon_\theta(x^{(t-1)}, t)}{d\delta_t} = \frac{\partial\epsilon_\theta(x^{(t-1)}, t)}{\partial x^{(t-1)}} \frac{dx^{(t-1)}}{d\delta_t} + \frac{\partial\epsilon_\theta(x^{(t-1)}, t)}{\partial t} \frac{dt}{d\delta_t}
\]
We find that the time-derivative term $\frac{\partial\epsilon_\theta(x^{(t-1)}, t)}{\partial t}$ is sufficient to distinguish between denoising path from high and low likelihood samples. Thus, we construct our pruning objective as:
\[
g(x^{(0)}) = \sum_{t=1}^T \left\| \frac{\partial\epsilon_\theta(x^{(t-1)}, t)}{\partial t} \right\|_2
\]
which is summing the curvature of the complete denoising timestep. A lower value of $g(x_0)$ indicates high-likelihood samples. The final objective of a sampled plan $\tau$ composing of segments $(x_1, x_2, \dots, x_{H})$, where $x_k = (s_{k-1}, \pi_{k-1}, a_{k-1}, s_{k})$, is calculated as:
\[
\prod_{k=1}^{H} \exp{\left(-g(x_k^{(0)})\right)}
\]
 Based on the cumulative score of all segments of a plan, we select top-M plans to move on to the next denoising timestep of the compositional sampling process.

 In this section, we want to understand the efficacy of the DDIM inversion based pruning objective.

\begin{rebuttal}
\subsection{Illustrative example: DDIM Inversion and OOD metrics}

\textbf{Experiment description:} We learn a 1D distribution of $x$ such that $[-1.0,-0.5]\cup[-0.1,0.2]\cup[0.6,1.0]$ is in-distribution~(ID) and remaining segments are out of distribution~(OOD) by construction. We learn a simple MLP score function to represent the diffusion model.

We draw clean samples uniformly from $[-1.0,1.0]$ and use DDIM inversion with the learned score function to noise them for 100 timesteps and then use 100 steps of DDIM denoising to reconstruct the clean samples back. Note that the original clean samples contain both ID and OOD while the reconstructed samples only contain ID.

We calculate the following metrics: 
\begin{enumerate}
    \item DDIM inversion metric: This is what is used in CDGS. The goal is to quantify the curvature of the inversion path. Smoother path means high-likelihood clean sample, while a path with abrupt direction changes mean low-likelihood clean samples. We only measure the cummulative curvature of the first 20\% inversion trajectory as, after that the path stabilizes as noisy latents come within in-distribution regions.
    \item Reconstruction metric: We calculate the error between the reconstructed sample and the clean sample. Note that this is after 100 steps of inversion followed by 100 denoising steps as shown in~\autoref{fig:ddim_inv_new}.
    \item Restoration Gap: This is another form of reconstruction metric but we do not need to inversion to obtain the noisy latents. We can sample any denoising timestep, add noise to the timestep using $x_t = \sqrt{\alpha_t}x_0 + \sqrt{1 - \alpha_t}\epsilon, \qquad \epsilon \in N(0,I)$ and then denoise $x_t$ from timestep $t$ to obtain reconstructed clean sample $\hat{x}^t_0$. Thus restoration gap can be calculated as: $\mathbb{E}_t [\hat{x}^t_0 - x_0]$. This can be repeated for multiple choice of timesteps.
    
\end{enumerate}

\begin{figure}[h]
    \centering
    \includegraphics[width=0.75\linewidth]{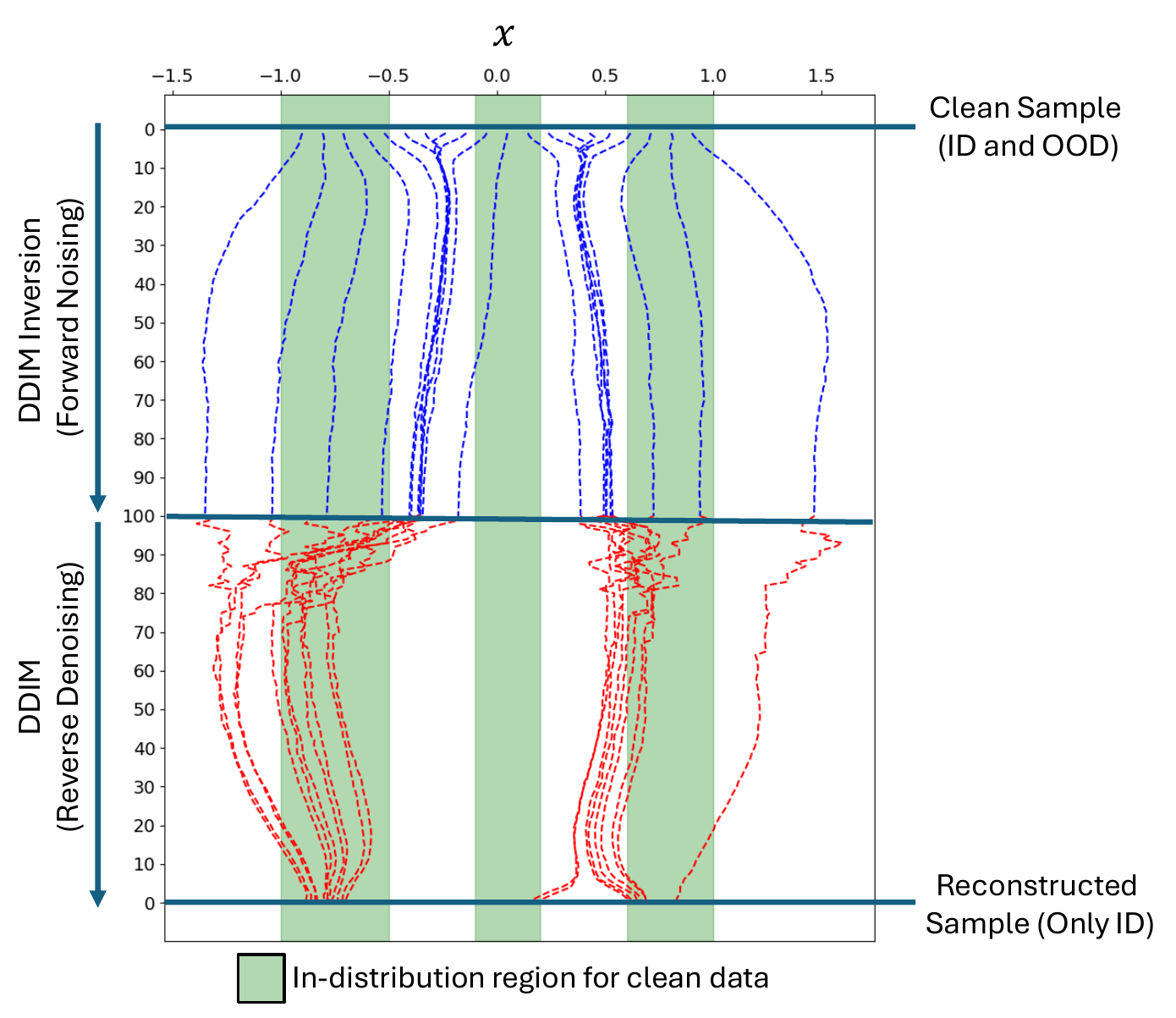}
    \caption{This plot contrasts the DDIM Inversion (forward noising, blue lines) with DDIM Denoising (reverse, red lines) on a 1D dataset where green areas mark the in-distribution (ID) regions. \textbf{Top.} (Inversion, $t=0 \to 100$) shows both ID and out-of-distribution (OOD) clean samples diffusing into noise using the learned score function. \textbf{Bottom.}(Denoising, $t=100 \to 0$) illustrates the learned model starting from noise and guiding all trajectories to reconstruct samples only within the valid ID regions, demonstrating how OOD paths are pulled back to the data manifold.}
    \label{fig:ddim_inv_new}
\end{figure}

\textbf{Advantages of the curvature-based approach over reconstruction-based alternatives for likelihood approximation:} We see two directions of improvement when using CDGS's curvature-based metric vs reconstruction-based alternatives:
\begin{enumerate}
    \item DDIM inversion only requires forward noising while reconstruction methods require both forward noising and denoising back.
    \item For distributions with disjoint modes (like the one considered for this experiment), it is not necessary that the reconstructed sample after noising and denoising will belong to the same mode as the original clean sample. This makes reconstruction-based metrics invalid or overly conservative, neglecting in-distribution segments. We show this in~\autoref{fig:ddim_metric_new} where the ID samples from middle segment after reconstruction belong to the left and right segments. While this increases the reconstruction error, the curvature metric can robustly handle this phenomenon. On the other hand, the restoration gap fails to give any meaningful signal.
\end{enumerate}

\begin{figure}[h]
    \centering
    \includegraphics[width=0.8\linewidth]{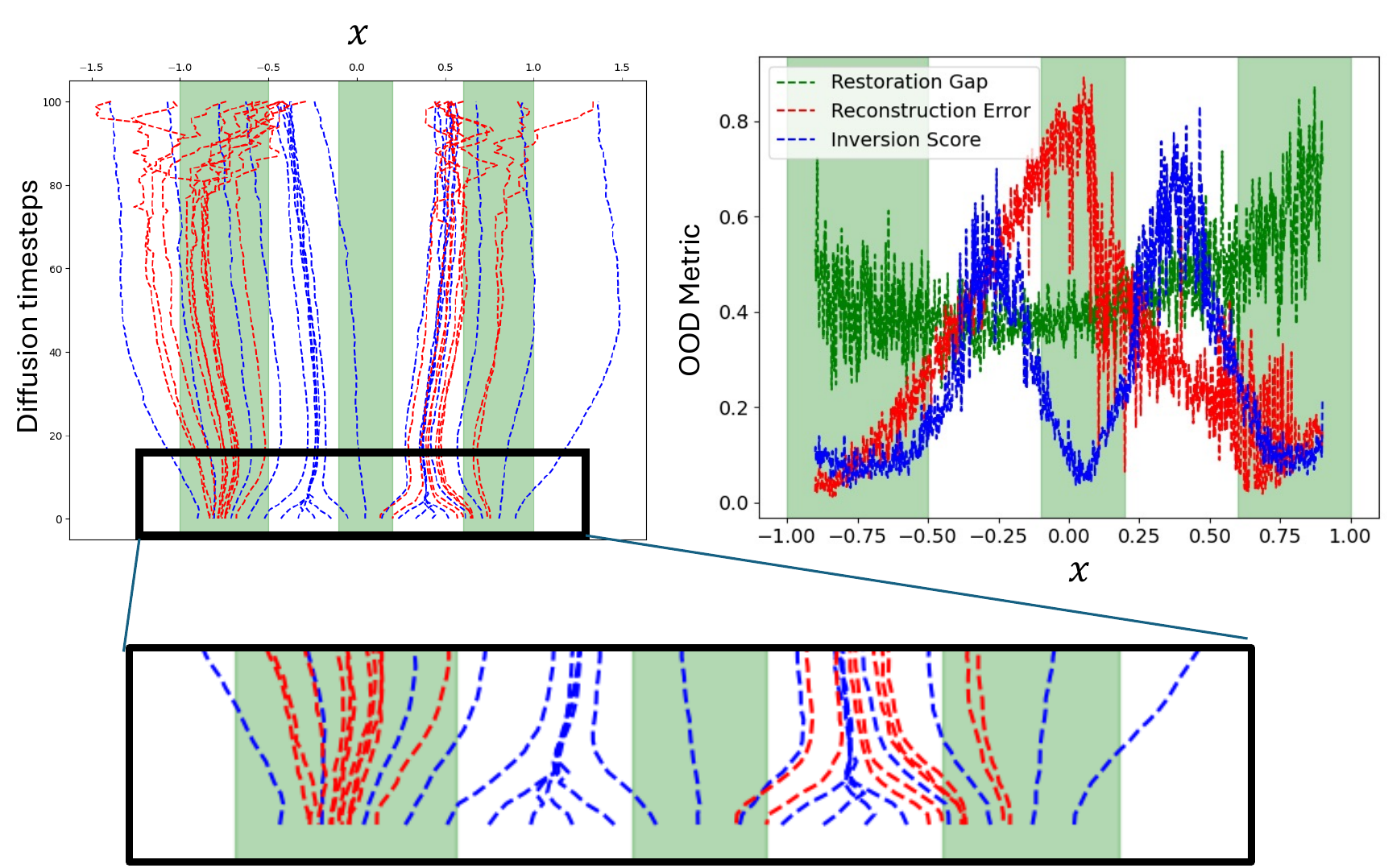}
    \caption{\textbf{Comparing DDIM trajectories and associated OOD metrics.} \textbf{Left} shows superimposed DDIM inversion (noising, blue) and denoising (reconstruction, red) paths. The blue lines show samples starting from both ID and OOD regions (e.g., the middle segment) being noised. The red lines show that all trajectories, when denoised, are guided back to the ID (green) regions. \textbf{Bottom} highlights the initial steps of the inversion (noising) paths. It illustrates that paths starting from OOD samples exhibit abrupt changes in noising directions, while paths starting in-distribution are smoother. \textbf{Right} compares OOD metrics (where a lower score is better). The Inversion Score (blue) accurately identifies the OOD and ID regions. The Reconstruction Error (red) is overly conservative, incorrectly flagging the middle segment as OOD. The Restoration Gap (green) provides no useful signal, failing to distinguish between ID and OOD regions.}
    \label{fig:ddim_metric_new}
\end{figure}
\end{rebuttal}

\section{Additional TAMP suite details}
\label{app:tasks}

We evaluate our framework on three task domains~(\texttt{hook reach}, \texttt{rearrangement push}, and \texttt{rearrangement memory}) with two tasks each. Each of the considered suites focuses on understanding long-horizon success of one particular skill. For example, \texttt{hook reach} is about the long-term effect of executing \texttt{hook}, while \texttt{rearrangement push} focuses on \texttt{push} and \texttt{rearrangement memory} is designed to confuse the TAMP framework that perform hierarchical planning with non goal-conditioned motion planners. Each task's challenge is directly proportional to the long-horizon action dependency required to complete it. For example, \texttt{pull} affects immediately if the next skill is \texttt{pick}. But \texttt{place} affects the next skill after executing one intermediate skill (like \texttt{pick}). Similarly, action dependency is after two skills for \texttt{rearrangement push} and \texttt{rearrangement memory} tasks. We describe all of such considered tasks below.

\begin{enumerate}
    \item \textbf{Hook Reach (Task 1):}
        \begin{itemize}
            \item \textbf{Scene:} Table with a rack, hook, and cube
            \item \textbf{Start:} Rack and hook are in workspace, cube is beyond workspace
            \item \textbf{Goal:} Pick up the cube
            \item \textbf{Action Skeleton:} pick(hook) $\rightarrow$ pull(cube, hook) $\rightarrow$ place(hook) $\rightarrow$ pick(cube)
        \end{itemize}
    
    \item \textbf{Hook Reach (Task 2):}
        \begin{itemize}
            \item \textbf{Scene:} Table with a rack, hook, and cube
            \item \textbf{Start:} Rack and hook are in workspace, cube is beyond workspace
            \item \textbf{Goal:} Place the cube on the rack
            \item \textbf{Action Skeleton:} pick(hook) $\rightarrow$ pull(cube, hook) $\rightarrow$ place(hook) $\rightarrow$ pick(cube) $\rightarrow$ place(cube, rack)
        \end{itemize}

    \begin{figure}[h]
        \centering
        \includegraphics[width=0.7\linewidth]{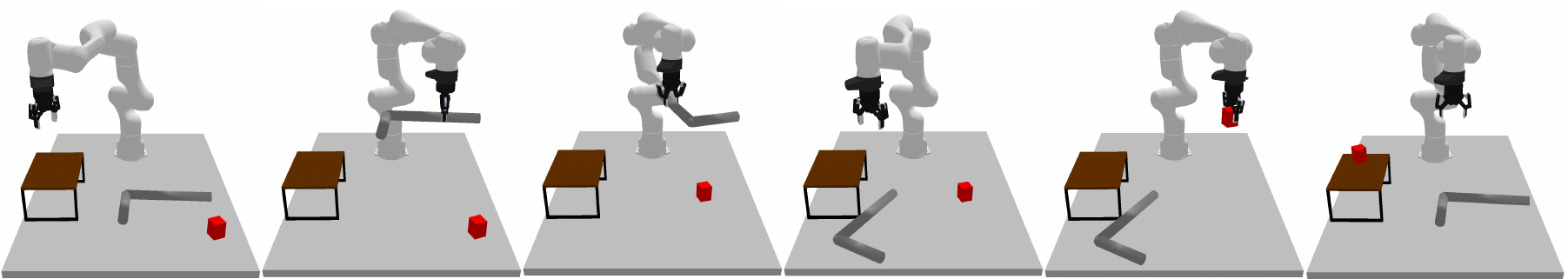}
        \caption{Hook Reach Task 2}
        \label{fig:HP_T2}
    \end{figure}
    
    \item \textbf{Rearrangement Push (Task 1):}
        \begin{itemize}
            \item \textbf{Scene:} Table with a hook, cube, and rack
            \item \textbf{Start:} Hook and cube are in workspace, rack is beyond workspace
            \item \textbf{Goal:} Position the cube under the rack
            \item \textbf{Action Skeleton:} pick(cube) $\rightarrow$ place(cube) $\rightarrow$ pick(hook) $\rightarrow$ push(cube, hook, rack)
        \end{itemize}
    
    \item \textbf{Rearrangement Push (Task 2):}
        \begin{itemize}
            \item \textbf{Scene:} Table with a hook, cube, and rack
            \item \textbf{Start:} Hook is in workspace, cube and rack are beyond workspace
            \item \textbf{Goal:} Position the cube under the rack
            \item \textbf{Action Skeleton:} pick(hook) $\rightarrow$ pull(cube, hook) $\rightarrow$ place(hook) $\rightarrow$ pick(cube) $\rightarrow$ place(cube) $\rightarrow$ pick(hook) $\rightarrow$ push(cube, hook, rack)
        \end{itemize}

    \begin{figure}[h]
        \centering
        \includegraphics[width=\linewidth]{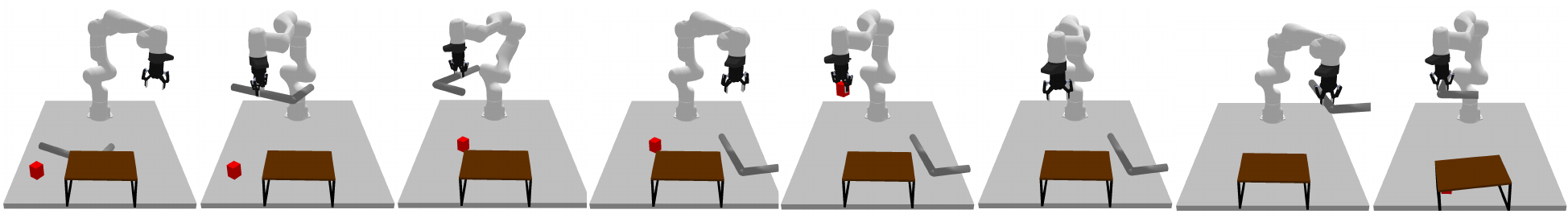}
        \caption{Rearrangement Push Task 2}
        \label{fig:RP_T2}
    \end{figure}
    
    \item \textbf{Rearrangement Memory (Task 1):}
        \begin{itemize}
            \item \textbf{Scene:} Table with a hook, red cube, and blue cube
            \item \textbf{Start:} All objects (hook, red cube, blue cube) are in workspace
            \item \textbf{Goal:} Put the red cube where the blue cube is
            \item \textbf{Action Skeleton:} pick(blue\_cube) $\rightarrow$ place(blue\_cube) $\rightarrow$ pick(red\_cube) $\rightarrow$ place(red\_cube)
        \end{itemize}
    
    \item \textbf{Rearrangement Memory (Task 2):}
        \begin{itemize}
            \item \textbf{Scene:} Table with a hook,red cube, and blue cube
            \item \textbf{Start:} Hook and blue cube are in workspace, red cube is beyond workspace
            \item \textbf{Goal:} Put the red cube where the blue cube is
            \item \textbf{Action Skeleton:} pick(hook) $\rightarrow$ pull(red\_cube, hook) $\rightarrow$ place(hook) $\rightarrow$ pick(blue\_cube) $\rightarrow$ place(blue\_cube) $\rightarrow$ pick(red\_cube) $\rightarrow$ place(red\_cube)
        \end{itemize}

    \begin{figure}[h]
        \centering
        \includegraphics[width=\linewidth]{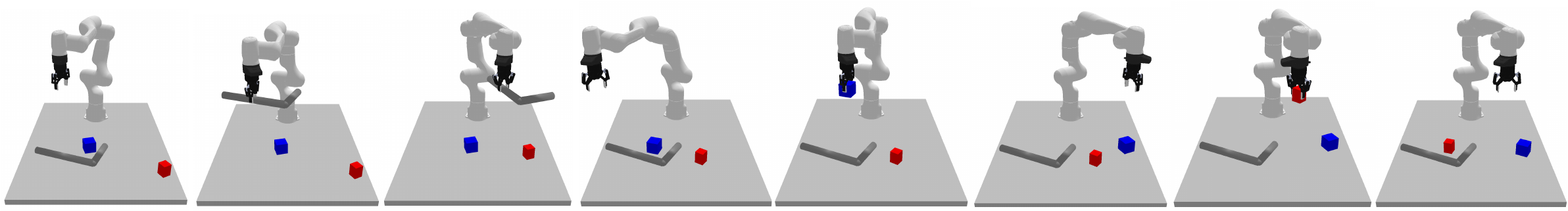}
        \caption{Rearrangement Memory Task 2}
        \label{fig:RM_T2}
    \end{figure}
\end{enumerate}

\subsection{Skill Structure}
\label{app:data-collection}

We consider a finite set of parameterized skills in our skill library. The parameterization, data collection, and training method for each of the skills is described as follows:

\begin{enumerate}
    \item \texttt{Pick}:~Gripper picks up an object from the table and the parameters contain 4-DoF pose in the object's frame of reference $(x, y, z, \theta )$. 
    \item \texttt{Place}:~Gripper places an object at the target location and parameters contain 4-DoF pose in the place target's frame of reference $(x, y, z, \theta )$. This skill requires specifying two set of parameters, the target pose and the target object (e.g. hook, table).
    \item \texttt{Push}:~Gripper uses the grasped object to push away another object. The skill is motivated from prior work~\cite{mishra2023generative, agia2022taps} where a hook object is used to \texttt{Push} blocks. The parameters of this skill are $(x, y, r, \theta)$ such that the hook is placed at the $(x, y)$ position on the table and pushed by a distance $r$ in the radial direction $\theta$ w.r.t. the origin of the manipulator.
    \item \texttt{Pull}:~Gripper uses the grasped object to pull another object inwards. The skill is also motivated from prior work~\cite{mishra2023generative, agia2022taps} where a hook object is used to \texttt{Pull} blocks. The parameters of this skill are $(x, y, r, \theta)$ such that the hook is placed at the $(x, y)$ position on the table and pulled by a distance $r$ in the radial direction $\theta$ w.r.t. the origin of the manipulator.
\end{enumerate}

\subsection{State Space of the Unified Skill Transition Model}

\ours{} assumes access to 6D object poses. In practice, we construct the system state as a concatenated vector of poses of objects present in the scenario. We use a fixed object order ([robot, rack, hook, cube1, cube2, …]), passing zero-vectors for absent objects, consistent across all baselines for the experiment.

\section{Training and Sampling: More details on TAMP experiments}
\label{app:hparams}

\subsection{\ours{}: Unified Score Model Training}

For our TAMP suite, we collect $10000$ random skill transition demonstrations for each skill by rolling out random policies in the environment. This ensures enough diversity in the system transitions in the training data. As shown in~\autoref{fig:network_arch}, we use a mixture-of-experts (MOE) model where we use $N$ feedforward MOE layers. Each layer has a gating network and $M$ experts, where diffusion timestep information is used through an adaptive layer normalization~(AdaLN) layer. The outputs from each expert are merged using the predicted gating softmax weights to get the final score of the noisy transition tuple. For OGbnch~\cite{park2024ogbench}, we just use the datasets provided by them: \url{https://github.com/seohongpark/ogbench}

\begin{figure}[h]
    \centering
    \includegraphics[width=1\linewidth]{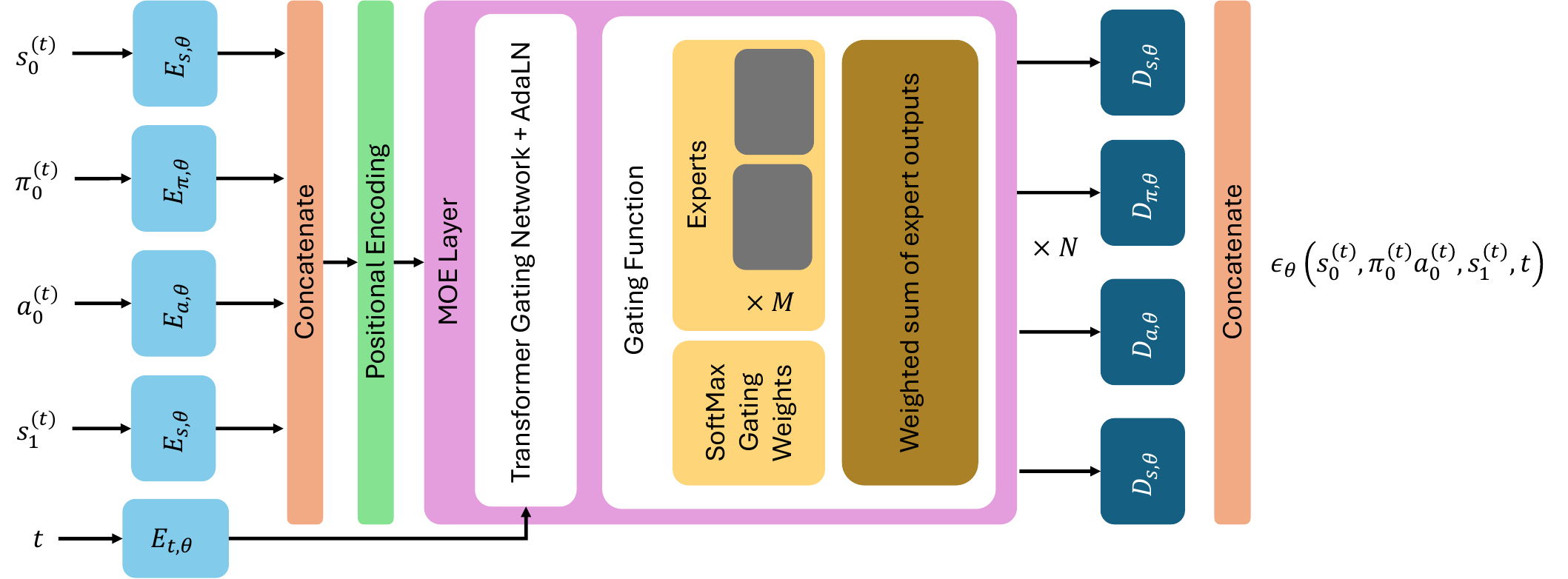}
    \caption{Network architecture for the score function}
    \label{fig:network_arch}
\end{figure}

We particularly use \url{https://huggingface.co/docs/diffusers/en/index} library to deploy training and sampling. We provide the training hyperparameters of our setup below:
\begin{table}[h]
    \centering
    \caption{Training setup hyperparameters for \ours{}}
    \label{tab:gtamp_training}
    \begin{tabular}{lc}
        \toprule
        \textbf{Hyperparameter} & \textbf{Value} \\
        \midrule
        Num. MOE layers & $3$ \\
        Num. Experts per layer & $6$ \\
        Encoder output dim & $256$\\
        Gating network Transformer num. heads & $4$\\
        Hidden-dims & $256$\\
        Optimizer & \texttt{torch.AdamW}\\
        Learning rate & $1e-4$\\
        Positional Encoding & \texttt{sinusoidal}\\
        Num. Training Steps & $1e6$\\
        Num Diffusion timesteps & $500$ \\
        Diffusion $\beta$ schedule & \texttt{cosine} \\
        Prediction type & \texttt{epsilon} \\
        \bottomrule
    \end{tabular}
\end{table}

\textbf{Effect of training data coverage.} If we consider an ``ideal" score function and a perfect representation of the system transition distributions, a solution exists if there is an overlap between the pre-condition and effect of two chosen skills that are required to solve the plan. If such an overlapping segment does not exist, \ours{} will not be able to complete the plan. Hence, the training data for each skill must be diverse enough to ensure that the overlap exists. Also, it is worth noting that we use separate dataloaders for all skills to ensure equal distribution of skills in training batches and thus equal preference when sampling.

\subsection{CDGS: Sampling Strategy}
For the main denoising loop, we use $T$ denoising timesteps and start with a initial batch size of $B$. For a plan of horizon $H$, we perform the compositional score computation and iterative resampling with $U(t)$ number of resampling steps at every denoising timestep $t$. We devise an adaptive strategy where we apply 
\begin{enumerate}
    \item no pruning for the first few denoising steps until $T_e = k_eT$. We call this as exploration phase. We keep the number of resampling iterations to low during this phase.
    \item pruning starts from $T_e = k_eT$ and is done until $T_p = k_pT$. During this, at each denoising timestep, we do some resampling iterations $U(t)$ and then select top-$K$ elites based on the pruning metric.
    \item once we have potentially high-quality globally coherent sequences of local modes after a few steps of pruning, we start increasing resampling iterations. This allows us to align the local plans more closely with the optimal mode sequences.
\end{enumerate}
We show the value of each hyperparameter in~\autoref{tab:gtamp_sampling}.

\begin{table}[h]
    \centering
    \caption{TAMP suite experiments: Sampling setup hyperparameters for \ours{}}
    \label{tab:gtamp_sampling}
    \begin{tabular}{lc}
        \toprule
        \textbf{Hyperparameter} & \textbf{Value} \\
        \midrule
        Denoising timesteps $T$ & $10$ \\
        Batch size $B$ & $100$ for $H=7$ and $50$ for $H=4 \& 5$ \\
        Resampling schedule $U(t)$ & $\frac{T - t + 1}{T}(U_{T})$\\
        Maximum resampling steps $U_T$ & $50$ for $H=7$ and $40$ for $H=4 \& 5$\\
        Exploration ends at $k_e$ & $0.7$\\
        Pruning ends at $k_p$ & $0.3$\\
        Top-$K$ pruning selection & $0.2 \times B$\\
        Pruning objective calculated with $P$ DDIM inversion steps & $0.4 \times T$ \\
        \bottomrule
    \end{tabular}
\end{table}

Thus for a plan of horizon $H$, the total number of function evaluation~(NFE) comes to be:
\[
NFE = \underbrace{U_T \times \frac{T(T+1)}{2}}_{\text{Main Denoising Loop}} + \underbrace{(k_e - k_p)T \times P}_{\text{Pruning phase}}
\]
Since using a single model allows batch operations of converting the $B$ plans of horizon $H$ into a single batched model evaluation with $B \times H$ short transitions.

\subsection{\ours{}: Runtime and evaluation}
We observe the inference time of \ours{} to be $0.5 \times H$ sec (\emph{linear} with $H$) on an Nvidia L40s GPU where $H$ is the plan length. For success metrics, we consider a task success according to the following: (1) \textbf{Hook Reach:} the cube is on rack in a stable position (2) \textbf{Rearrangement Push:}  $\geq50\%$ of the cube is under the rack and (3) \textbf{Rearrangement Memory:} cube within 0.05 m of target positions. 

\subsection{STAP~\cite{agia2022taps}}
For STAP, we use their policies, critics, and dynamics models trained with their inverse reinforcement learning pipeline (text2motion ~\cite{lin2023text2motion}) available at \url{https://github.com/agiachris/STAP}. For Rearrangement Push, we modify the criteria of the \texttt{Under} predicate such that $>=50\%$ of the cube must be under the rack to be successful. We train a new model using STAP's code for \texttt{Push} and use their pre-trained models for the other skills.

\subsubsection{Task Planning}
STAP by itself is only a motion planner. In order to solve full TAMP problems, it must be integrated with an external task planner. Symbolically-feasible skill sequences found by the task planner are evaluated and ranked by STAP for geometric feasibility. For our experiments, we use a BFS-based symbolic planner that searches through a hand-designed PDDL domain. 

To better reflect practical considerations, we design the PDDL domain for each task so that provided geometric information is minimized while ensuring that the correct task plan can always be found. We modify the BFS algorithm so that it can revisit previously visited states as the hidden geometric predicates may be different despite the same symbolic predicates. 



\textbf{Remark on Rearrangement Memory task.} There are two particular characteristics required in a TAMP to solve Rearrangement Memory task:
\begin{enumerate}
    \item The symbolic planner must understand which particular symbolic state will satisfy the goal condition. Since most required skills are \texttt{pick} and \texttt{place}, the symbolic effect of all \texttt{place} actions are same. As the exact goal position is not embed in the symbolic states, it is not possible for a na\"ive task planner to solve for a skill sequence.
    \item The task planner can give many feasible solutions that the motion planner must evaluate to find the final task and motion plan. This requires goal-conditioned planners. Since STAP uses Q-function based value estimates to evaluate plans, we find that it struggles with the task as it is not a goal-conditioned method.
\end{enumerate}


\subsubsection{CEM Sampling}
The STAP baselines use a CEM-based sampling algorithm. An initial prior for the actions is sampled for the start state, and then optimized using the value and dynamics models with CEM-optimization. The only difference between Random CEM and STAP CEM is that Random CEM samples the prior from a uniform distribution? (double-check) while STAP samples the prior from its learned policy models

To make a fair comparison between \ours{}'s diffusion-based sampling and STAP's CEM-based sampling, we match the sampling budget based on the number of function evaluations. Since our unified model serves the same purpose as STAP's   policy, value, and dynamics models, we consider evaluating one set of STAP's policy, value, and dynamics models for a skill to be one function evaluation. For STAP, the CEM runs num iterations of sampling for $N * batch_size samples$. Thus, we match the number of sampling iterations and batch size. The exact budgets for each task are given below.

\begin{table}[htbp]
    \centering
    \caption{CEM-sampling parameters for STAP}
    \label{tab:experiment_params}
    \begin{tabular}{lccccc}
        \toprule
        \textbf{Task} & \textbf{Samples} & \textbf{Iterations} & \textbf{Elites} & \textbf{Total NFE} & \textbf{\ours{} NFE}\\
        \midrule
        Hook Reach 1 & 40 & 132 & 16 & 5280 & 2300\\
        Hook Reach 2 & 50 & 165 & 20 & 8250 & 2300\\
        Rearrangement Push 1 & 50 & 165 & 20 & 8250 & 2300\\
        Rearrangement Push 2 & 70 & 336 & 28 & 23520 & 2850\\
        Rearrangement Memory 1 & 40 & 132 & 16 & 5280 & 2300\\
        Rearrangement Memory 2 & 70 & 336 & 28 & 23520 & 2850\\
        \bottomrule
    \end{tabular}
\end{table}

\subsubsection{Uncertainty Quantification}
The LLM Planner in text2motion \cite{lin2023text2motion} can sometimes generate symbolically invalid actions i.e. (place(cube) when nothing is in hand), which are out-of-distribution for the learned models. text2motion uses a simple ensemble-based OOD detection method (detailed in appendix A.2 of their paper) to filter out symbolically-invalid actions. We use this for all text2motion baselines. For completeness, we also include this in STAP's baselines as \textbf{STAP CEM + UQ}, but it does not make any significant improvements.



\section{Additional Ablations}
\label{app:ablations}

\subsection{\ours{}: A better prior for task-level trajectory sampling}

The proposed method constructs a task-level distribution from skill-level distribution given the current state, the intended goal state and the planning horizon. Specifically, \ours{} finds a sequence of modes with overlapping pre-condition and effects by systematic exploration and pruning. While this does not always ensure that the plan is symbolically-geometrically feasible, we observe that choosing the top two plans and expanding the BFS tree with system rollouts leads to higher success rates. As shown in~\autoref{tab:gtamp_bfs}, the \ours{} (BFS-2) proves to be an upper bound of our approach. This points out that \ours{} constructs meaningful task-level distribution with correct task plans. 

\begin{table}[h]
    \centering
    \spacing{-1em}
    \caption{The success rate of the proposed \ours{} algorithm is shown and compared with a variant that performs BFS with the top-2 skill chains at every step and uses system dynamics to rollout. All results are calculated from $50$ trials for each task. }
    \begin{tabular}{|l|c|c|c|c|c|c|}
    \hline
       \multirow{2}{*}{} & \multicolumn{2}{c|}{Hook Reach} & \multicolumn{2}{c|}{Rearrangement Push} & \multicolumn{2}{c|}{Rearrangement Memory}\\
       \cline{2-7}
       & Task 1 & Task 2 & Task 1 & Task 2 & Task 1 & Task 2 \\
       \hline         Task Length & 4 & 5 & 4 & 7 & 4 & 7\\
       \hline 
        \hline 
       \multicolumn{7}{|c|}{Full Generative TAMP (no PDDL, skill-level data only)} \\
       \hline
       \ours {} (ours) & 0.64 & 0.58 & 0.84 & 0.48 & 0.42 & 0.18\\
        \hline 
        \hline 
       \multicolumn{7}{|c|}{Full Generative TAMP (no PDDL, skill-level data only) + Rollout with system dynamics} \\
       \hline
        \ours {} (BFS-2) & 0.72 & 0.64 & 0.90 & 0.62 & 0.48 & 0.22\\
         \hline 
    \end{tabular}
    \label{tab:gtamp_bfs}
    \spacing{-1em}
\end{table}

\subsection{Analyzing scaling for individual tasks}

We analyze how varying the batch size $B$ and the number of resampling iterations $U$ affects overall planning performance across
\begin{figure}[h]
    \centering
    \includegraphics[width=0.8\linewidth]{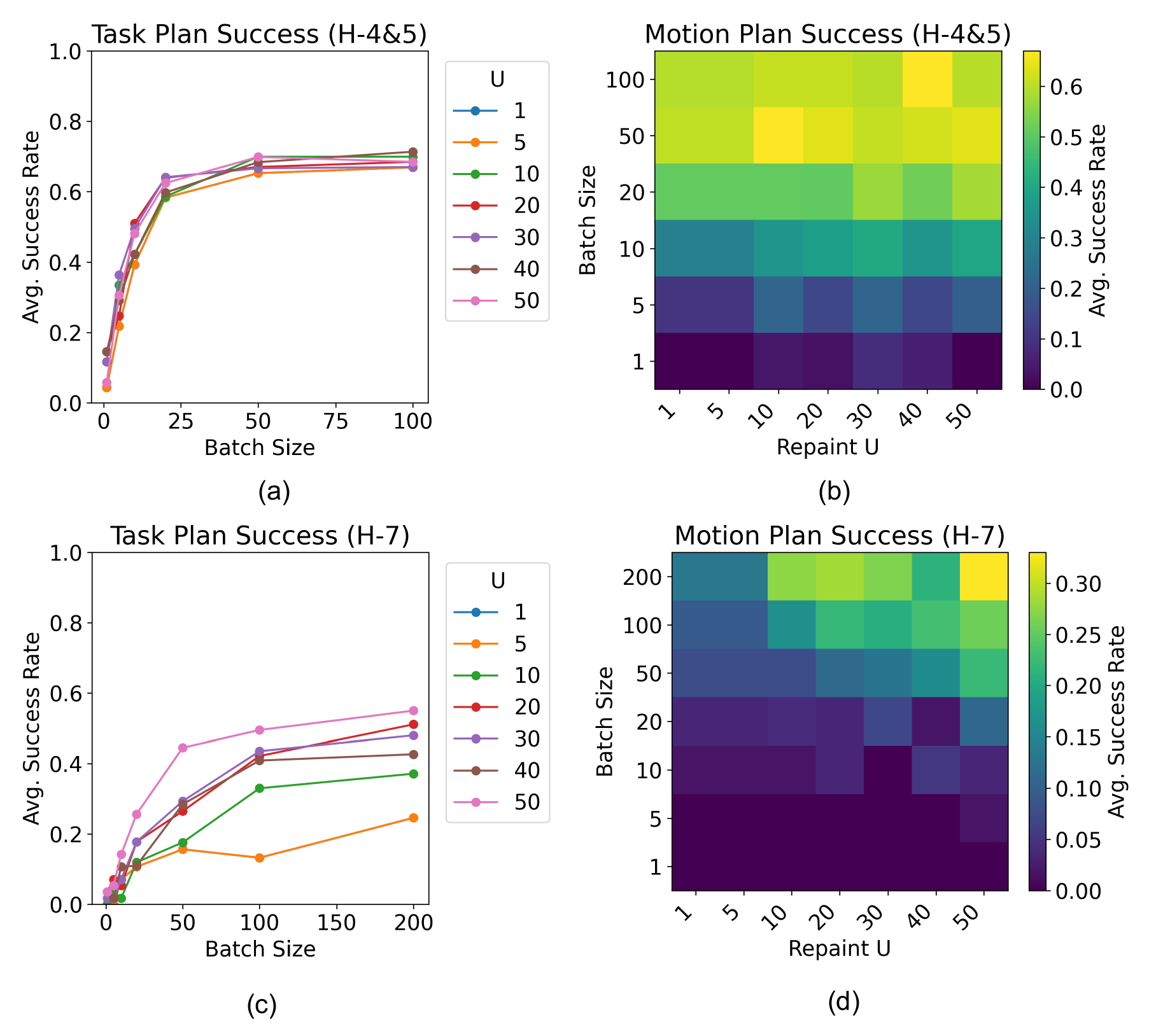}
    \caption{We show the effect of scaling $B$ and $U$ on the overall task planning and motion planning success of \ours{} for shorter tasks ($H = 4\&5$) in (a,b) and longer tasks ($H = 7$) in (c,d) }
    \label{fig:scaling_all_tasks}
\end{figure}all long‑horizon tasks of horizon ($H$) $4\&5$ (Hook Reach Task 1 and Task 2; Rearrangement Push Task 1; Rearrangement Memory Task 1) and longer $H=7$ (Rearrangement Push Task 2; Rearrangement Memory Task 2) in~\autoref{fig:scaling_all_tasks}.
As shown in~\autoref{fig:scaling_all_tasks}(a, c), increasing $B$ yields a clear, monotonic rise in task‑planning success: larger candidate sets diversify the search over the task‑level distribution and enable the pruning stage to more reliably identify viable skill sequences.  Motion‑planning success exhibits a similar trend in~\autoref{fig:scaling_all_tasks}(b, d), demonstrating that a more diverse initial sample pool benefits the motion-planning optimization as well. We can also see\autoref{fig:scaling_all_tasks}(b, d) that at lower batch size, increasing the number of resampling steps yields only marginal improvements: without pruning, repeated denoising can still suffer from mode‑averaging local minima, where incorrect skill sequences become self‑reinforcing. It is only when resampling is coupled with pruning that results in better task planning as well as permit bidirectional “message‑passing” of information between the start and goal states—compensating for temporal misalignments at skill (pre-condition and effect) intersection—and thereby unlock significant gains in both task and motion success rates.

\section{Hardware Setup}
\label{app:hardware}

\begin{figure}[h]
    \centering
    \includegraphics[width=0.5\linewidth]{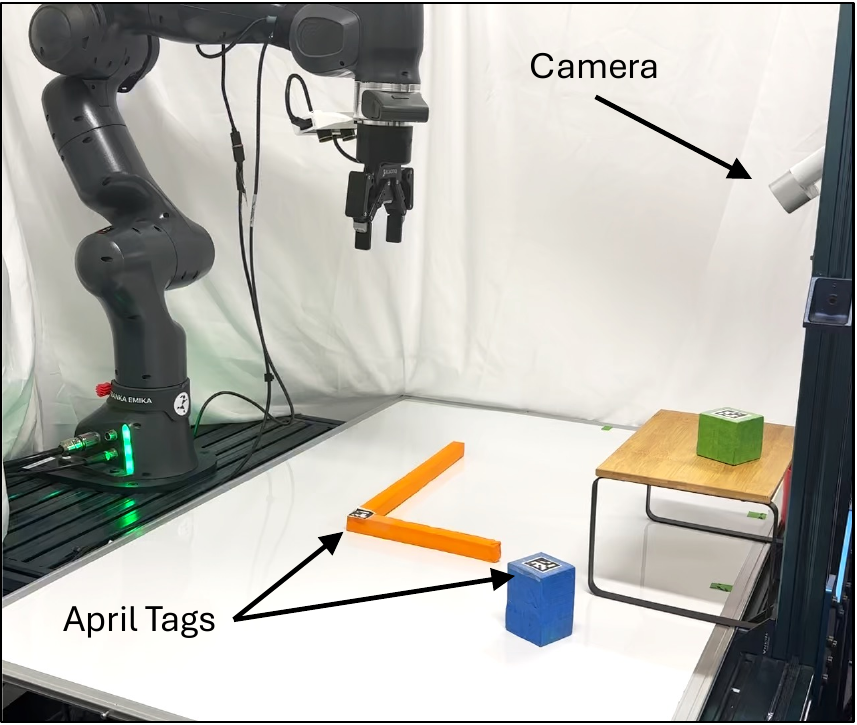}
    \caption{Hardware setup}
    \label{fig:hardware}
\end{figure}

The experimental setup, illustrated in~\autoref{fig:hardware}, consists of the same Franka Panda robot arm, several blocks, a rack, and a hook, observed by an Azure Kinect camera. The camera is mounted in an inclined front‑view configuration. AprilTag~\cite{apriltag} (\url{https://github.com/fabrizioschiano/apriltag2}) markers are used for SE(3) pose detection. We employ Deoxys~\cite{zhu2022viola} (\url{https://github.com/UT-Austin-RPL/deoxys_control}) for control. After obtaining the SE(3) poses of all the objects: (1) we construct the same environment in simulation, (2) deploy our algorithm in simulation, and (3) execute the planned action in real environment and finally replan.

\section{LLM and VLM Prompting}
\label{app:llm_prompting}

\subsection{LLM Prompting}
To make the fairest comparison, we use the same prompt style and in-context examples as text2motion\cite{lin2023text2motion}, which can be found in Appendix B.2. of their paper. We find that having the model generate multiple candidate task-plans during the shooting phase is critical to task performance, so we add a minimal system prompt to make the LLM instruction-following explicit. An example of a full prompt for \textbf{LLM-T2M, $n=1$} for \texttt{hook Reach} Task 1 is shown below.


\begin{figure}[h]
\centering
\begin{tcolorbox}[
    colback=gray!5,
    colframe=gray!50!black,
    title={\textbf{User Prompt}},
    fonttitle=\bfseries,
    boxrule=0.5pt,
    arc=0pt,
    boxsep=5pt,
    left=10pt,
    right=10pt
]
Respond directly in the format specified in the output format section, following the instructions exactly for how many sequences to generate i.e. generate 5 sequences if asked for the "Top 5 robot action sequences".\\

Available primitives: ['pick(a)', 'place(a, b)', 'pull(a, hook)', 'push(a, hook, rack)']
Available predicates: ['on(a, b)', 'inhand(a)', 'under(a, b)']\\
Available scene objects: ['table', 'blue\_box', 'cyan\_box', 'hook', 'rack', 'red\_box', 'yellow\_box']\\
Object relationships: ['inhand(hook)', 'on(red\_box, rack)', 'on(yellow\_box, table)', 'on(blue\_box, table)', 'on(cyan\_box, rack)', 'on(rack, table)']\\
Human instruction: could you move all the boxes onto the rack?\\
Goal predicate set: [['on(yellow\_box, rack)', 'on(blue\_box, rack)']]\\
Top 1 robot action sequences: ['pull(yellow\_box, hook)', 'place(hook, table)', 'pick(yellow\_box)', 'place(yellow\_box, rack)', 'pick(blue\_box)', 'place(blue\_box, rack)']\\

Available scene objects: ['table', 'rack', 'hook', 'red\_box']\\
Object relationships: ['on(rack, table)', 'on(hook, table)', 'on(red\_box, table)']\\
Human instruction: How would you get the red box in the robot's hand?\\
Goal predicate set: [['inhand(red\_box)']]\\
Top 5 robot action sequences (python list of lists): 
\end{tcolorbox}
\end{figure}

\begin{figure}[h]
\centering
\begin{tcolorbox}[
    colback=blue!5,
    colframe=blue!50!black,
    title={\textbf{Model Completion}},
    fonttitle=\bfseries,
    boxrule=0.5pt,
    arc=0pt,
    boxsep=5pt,
    left=10pt,
    right=10pt
]
[
  ['pick(red\_box)'],
  ['push(red\_box, hook, rack)', 'pick(red\_box)'],
  ['pick(hook)', 'place(hook, table)', 'pick(red\_box)'],
  ['pick(hook)', 'place(hook, rack)', 'pick(red\_box)'],
  ['pick(hook)', 'place(hook, table)', 'push(red\_box, hook, rack)', 'pick(red\_box)']
]
\end{tcolorbox}
\end{figure}

Interestingly, even though many of the prompts contain partial-to-complete solutions in the in-context examples like the one above, we find that this does not guarantee the LLM will generate the correct plan. 

A key piece to text2motion's success for planning despite LLM's lack of geometric awareness is the alternation between shooting and greedy search. When the shooting motion planner fails to find a geometrically feasible motion plan from the 5 task plan candidates, the algorithm falls back to a say-can style greedy search that executes the next action deemed most feasible by a combination of the LLM task planner and the motion planner. This allows the LLM task planner to explore with geometric feedback, and is the reason why text2motion is able to solve \texttt{Rearrangement Push} Task 1. For our evaluations, we allow 10 tries, which is much more than the steps required to solve any of the tasks in our evaluation suites.

\subsection{VLM Prompting}
For VLM experiments, we modify the system prompt and insert scene images before the scene description. Below is an example of \texttt{Hook Reach} Task 1 with $n=11$ in-context examples.

\begin{figure}[h]
\centering
\begin{tcolorbox}[
    colback=gray!5,
    colframe=gray!50!black,
    title={\textbf{User Prompt}},
    fonttitle=\bfseries,
    boxrule=0.5pt,
    arc=0pt,
    boxsep=5pt,
    left=10pt,
    right=10pt,
]
Respond directly in the format specified in the output format section, following the instructions exactly for how many sequences to generate i.e. generate 5 sequences if asked for the "Top 5 robot action sequences". \textcolor{cyan}{Review the provided images carefully when constructing your plan.}

Available primitives: ['pick(a)', 'place(a, b)', 'pull(a, hook)', 'push(a, hook, rack)']
Available predicates: ['on(a, b)', 'inhand(a)', 'under(a, b)']\\
Available scene objects: ['table', 'hook', 'rack', 'yellow\_box', 'blue\_box', 'red\_box']\\
Object relationships: ['inhand(hook)', 'on(yellow\_box, table)', 'on(rack, table)', 'on(blue\_box, table)']\\
Human instruction: How would you push two of the boxes to be under the rack?\\
Goal predicate set: [['under(yellow\_box, rack)', 'under(blue\_box, rack)'], ['under(blue\_box, rack)', 'under(red\_box, rack)'], ['under(yellow\_box, rack)', 'under(red\_box, rack)']]\\
Top 1 robot action sequences: ['push(yellow\_box, hook, rack)', 'push(red\_box, hook, rack)']\\

Available scene objects: ['table', 'blue\_box', 'cyan\_box', 'hook', 'rack', 'red\_box', 'yellow\_box']\\
Object relationships: ['inhand(hook)', 'on(red\_box, rack)', 'on(yellow\_box, table)', 'on(blue\_box, table)', 'on(cyan\_box, rack)', 'on(rack, table)']\\
Human instruction: could you move all the boxes onto the rack?\\
Goal predicate set: [['on(yellow\_box, rack)', 'on(blue\_box, rack)']]\\
Top 1 robot action sequences: ['pull(yellow\_box, hook)', 'place(hook, table)', 'pick(yellow\_box)', 'place(yellow\_box, rack)', 'pick(blue\_box)', 'place(blue\_box, rack)']\\

Available scene objects: ['table', 'blue\_box', 'hook', 'rack', 'red\_box', 'yellow\_box']\\
Object relationships: ['on(hook, table)', 'on(red\_box, table)', 'on(blue\_box, table)', 'on(yellow\_box, rack)', 'on(rack, table)']\\
Human instruction: Move the ocean colored box to be under the rack and ensure the hook ends up on the table.\\
Goal predicate set: [['under(blue\_box, rack)']]\\
Top 1 robot action sequences: ['pick(red\_box)', 'place(red\_box, table)', 'pick(yellow\_box)', 'place(yellow\_box, rack)', 'pick(hook)', 'push(blue\_box, hook, rack)', 'place(hook, table)']\\

Available scene objects: ['table', 'cyan\_box', 'hook', 'red\_box', 'yellow\_box', 'rack', 'blue\_box']\\
Object relationships: ['on(hook, table)', 'on(red\_box, table)', 'on(blue\_box, table)', 'on(cyan\_box, table)', 'on(rack, table)', 'under(yellow\_box, rack)']\\
Human instruction: How would you get the cyan box under the rack and then ensure the hook is on the table?\\
Goal predicate set: [['under(cyan\_box, rack)', 'on(hook, table)']]\\
Top 1 robot action sequences: ['pick(blue\_box)', 'place(blue\_box, table)', 'pick(red\_box)', 'place(red\_box, table)', 'pick(hook)', 'push(cyan\_box, hook, rack)', 'place(hook, table)']\\
\end{tcolorbox}
\end{figure}

\begin{figure}[h]
\centering
\begin{tcolorbox}[
    colback=gray!5,
    colframe=gray!50!black,
    title={\textbf{User Prompt}},
    fonttitle=\bfseries,
    boxrule=0.5pt,
    arc=0pt,
    boxsep=5pt,
    left=10pt,
    right=10pt,
]
Available scene objects: ['table', 'cyan\_box', 'hook', 'blue\_box', 'rack', 'red\_box']\\
Object relationships: ['on(hook, table)', 'on(rack, table)', 'on(blue\_box, table)', 'on(cyan\_box, table)', 'on(red\_box, table)']\\
Human instruction: How would you push all the boxes under the rack?
Goal predicate set: [['under(blue\_box, rack)', 'under(cyan\_box, rack)', 'under(red\_box, rack)']]\\
Top 1 robot action sequences: ['pick(blue\_box)', 'place(blue\_box, table)', 'pick(hook)', 'push(cyan\_box, hook, rack)', 'place(hook, table)', 'pick(blue\_box)', 'place(blue\_box, table)', 'pick(hook)', 'push(blue\_box, hook, rack)', 'push(red\_box, hook, rack)']\\

Available scene objects: ['table', 'cyan\_box', 'hook', 'rack', 'red\_box', 'blue\_box']\\
Object relationships: ['on(hook, table)', 'on(cyan\_box, rack)', 'on(rack, table)', 'on(red\_box, table)', 'inhand(blue\_box)']\\
Human instruction: How would you set the red box to be the only box on the rack?\\
Goal predicate set: [['on(red\_box, rack)', 'on(blue\_box, table)', 'on(cyan\_box, table)']]\\
Top 1 robot action sequences: ['place(blue\_box, table)', 'pick(hook)', 'pull(red\_box, hook)', 'place(hook, table)', 'pick(red\_box)', 'place(red\_box, rack)', 'pick(cyan\_box)', 'place(cyan\_box, table)']\\

Available scene objects: ['table', 'cyan\_box', 'red\_box', 'hook', 'rack']\\
Object relationships: ['on(hook, table)', 'on(rack, table)', 'on(cyan\_box, rack)', 'on(red\_box, rack)']\\
Human instruction: put the hook on the rack and stack the cyan box above the rack - thanks\\
Goal predicate set: [['on(hook, rack)', 'on(cyan\_box, rack)']]\\
Top 1 robot action sequences: ['pick(hook)', 'pull(cyan\_box, hook)', 'place(hook, rack)', 'pick(cyan\_box)', 'place(cyan\_box, rack)']\\

Available scene objects: ['table', 'cyan\_box', 'hook', 'rack', 'red\_box', 'blue\_box']\\
Object relationships: ['on(hook, table)', 'on(blue\_box, rack)', 'on(cyan\_box, table)', 'on(red\_box, table)', 'on(rack, table)']\\
Human instruction: Move the warm colored box to be underneath the rack.\\
Goal predicate set: [['under(red\_box, rack)']]\\
Top 1 robot action sequences: ['pick(blue\_box)', 'place(blue\_box, table)', 'pick(red\_box)', 'place(red\_box, table)', 'pick(hook)', 'push(red\_box, hook, rack)']\\

Available scene objects: ['table', 'blue\_box', 'red\_box', 'hook', 'rack', 'yellow\_box']\\
Object relationships: ['on(hook, table)', 'on(blue\_box, table)', 'on(rack, table)', 'on(red\_box, table)', 'on(yellow\_box, table)']\\
Human instruction: situate an odd number greater than 1 of the boxes above the rack\\
Goal predicate set: [['on(blue\_box, rack)', 'on(red\_box, rack)', 'on(yellow\_box, rack)']]\\
Top 1 robot action sequences: ['pick(hook)', 'pull(blue\_box, hook)', 'place(hook, table)', 'pick(blue\_box)', 'place(blue\_box, rack)', 'pick(red\_box)', 'place(red\_box, rack)', 'pick(yellow\_box)', 'place(yellow\_box, rack)']\\
\end{tcolorbox}
\end{figure}

\begin{figure}[h]
\centering
\begin{tcolorbox}[
    colback=gray!5,
    colframe=gray!50!black,
    title={\textbf{User Prompt}},
    fonttitle=\bfseries,
    boxrule=0.5pt,
    arc=0pt,
    boxsep=5pt,
    left=10pt,
    right=10pt,
]

continued... \\

Available scene objects: ['table', 'cyan\_box', 'hook', 'yellow\_box', 'blue\_box', 'rack']\\
Object relationships: ['on(hook, table)', 'on(yellow\_box, rack)', 'on(rack, table)', 'on(cyan\_box, rack)']\\
Human instruction: set the hook on the rack and stack the yellow box onto the table and set the cyan box on the rack\\
Goal predicate set: [['on(hook, rack)', 'on(yellow\_box, table)', 'on(cyan\_box, rack)']]\\
Top 1 robot action sequences: ['pick(yellow\_box)', 'place(yellow\_box, table)', 'pick(hook)', 'pull(yellow\_box, hook)', 'place(hook, table)']\\

Available scene objects: ['table', 'rack', 'hook', 'cyan\_box', 'yellow\_box', 'red\_box']\\
Object relationships: ['on(yellow\_box, table)', 'on(rack, table)', 'on(cyan\_box, table)', 'on(hook, table)', 'on(red\_box, rack)']\\
Human instruction: Pick up any box.\\
Goal predicate set: [['inhand(yellow\_box)'], ['inhand(cyan\_box)']]
Top 1 robot action sequences: ['pick(yellow\_box)']\\

        \includegraphics[width=0.3\textwidth]{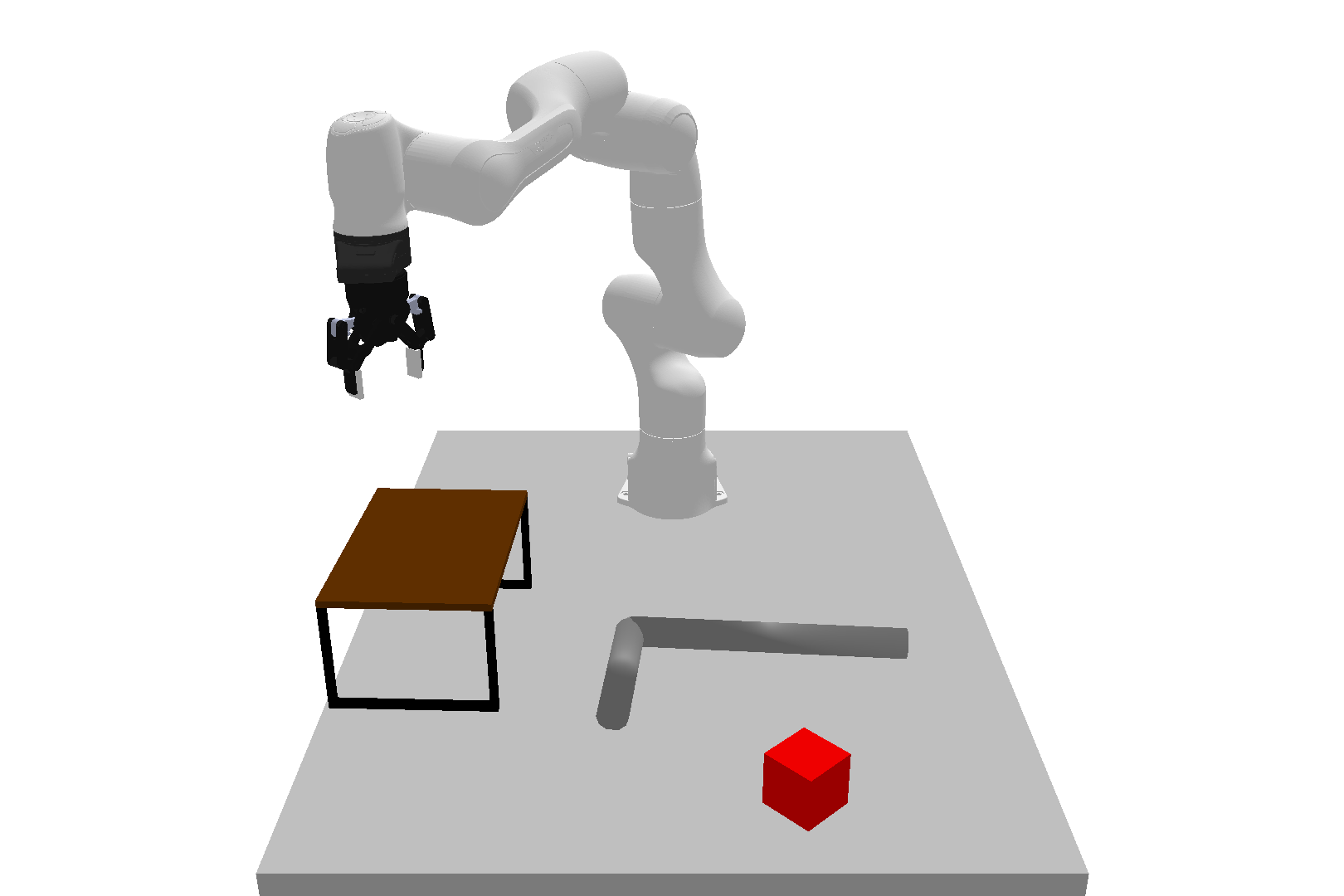}
        \label{fig:cnn}
    \hfill
        \includegraphics[width=0.3\textwidth]{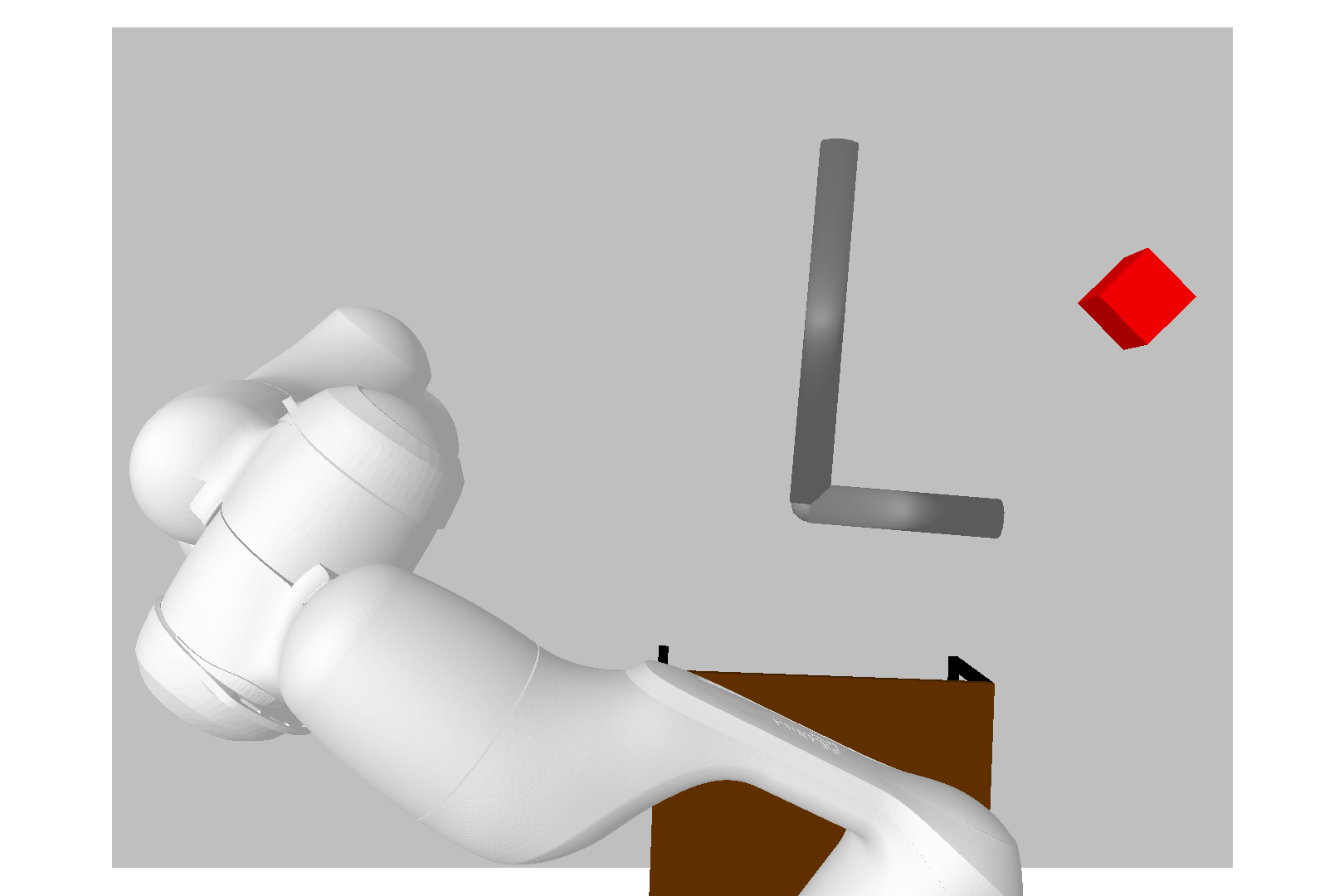}
        \label{fig:rnn}
    \hfill
        \includegraphics[width=0.3\textwidth]{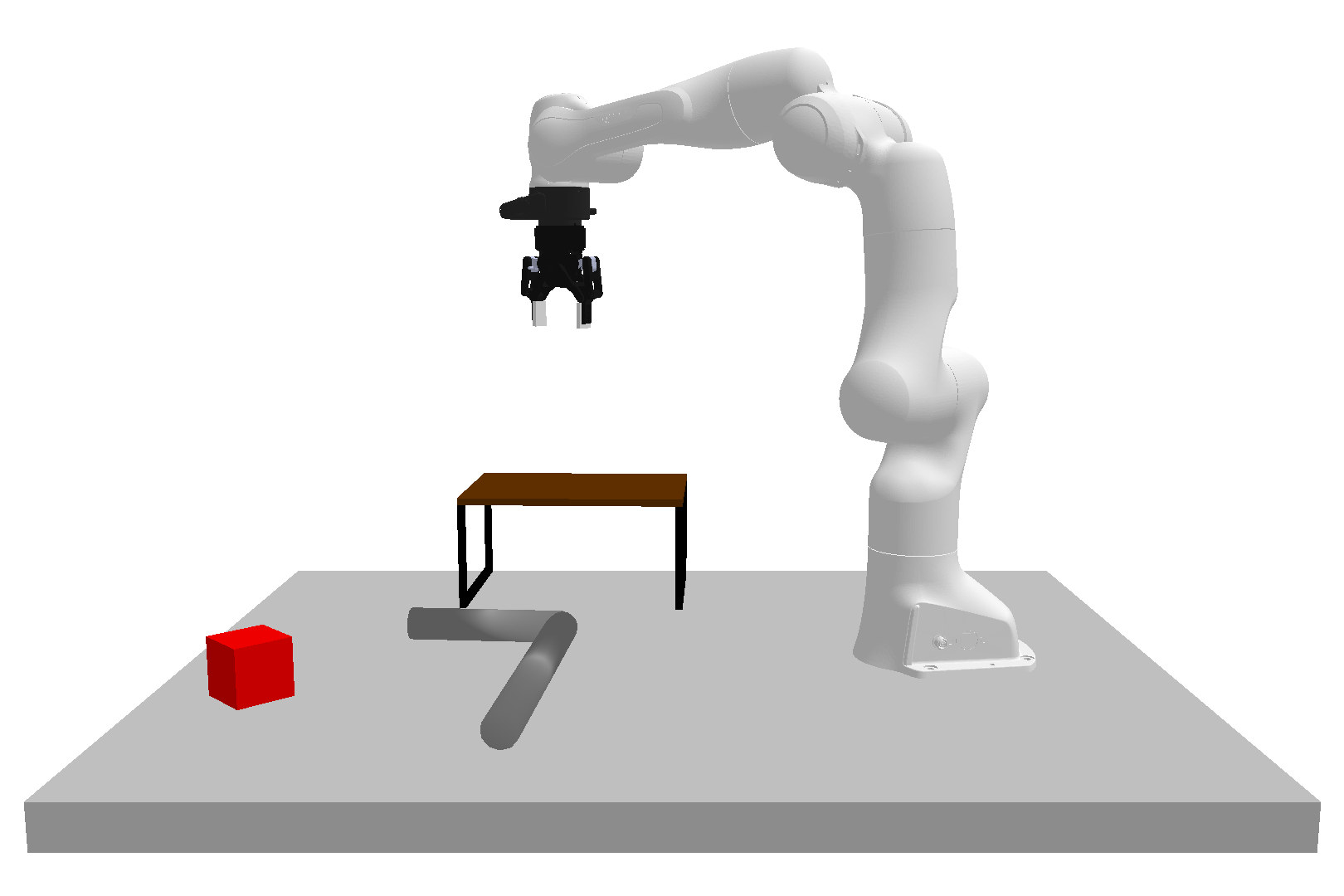}
        \label{fig:transformer}

Available scene objects: ['table', 'rack', 'hook', 'red\_box']\\
Object relationships: ['on(rack, table)', 'on(hook, table)', 'on(red\_box, table)']\\
Human instruction: How would you get the red box in the robot's hand?
Goal predicate set: [['inhand(red\_box)']]\\
Top 5 robot action sequences (python list of lists): 
\end{tcolorbox}
\end{figure}

\begin{figure}[h]
\centering
\begin{tcolorbox}[
    colback=blue!5,
    colframe=blue!50!black,
    title={\textbf{Model Completion}},
    fonttitle=\bfseries,
    boxrule=0.5pt,
    arc=0pt,
    boxsep=5pt,
    left=10pt,
    right=10pt
]
[
  ['pick(red\_box)'],
  ['pick(hook)', 'place(hook, table)', 'pick(red\_box)'],
  ['pick(rack)', 'place(rack, table)', 'pick(red\_box)'],
  ['pick(hook)', 'place(hook, table)', 'pick(rack)', 'place(rack, table)', 'pick(red\_box)'],
  ['pick(red\_box)', 'place(red\_box, table)', 'pick(red\_box)']
]
\end{tcolorbox}
\end{figure}

Interestingly, we find that including images in the prompt degrades the performance. 

\section{Hyperparameters for Image Generation}
\label{app:sec:image-code}

\begin{table}[h]
    \centering
    \caption{Sampling setup hyperparameters for panorama generation experiments}
    \label{tab:image_sampling}
    \begin{tabular}{lc}
        \toprule
        \textbf{Hyperparameter} & \textbf{Value} \\
        \midrule
        Denoising timesteps $T$ & $50$ \\
        Batch size $B$ & $10$ \\
        Composition weights $\gamma_{1:H}$ & $0.5$ \\
        Resampling schedule $U(t)$ & $\frac{T - t + 1}{T}(U_{T})$\\
        Maximum resampling steps $U_T$ & $10$\\
        Exploration ends at $k_e$ & $0.2$\\
        Pruning ends at $k_p$ & $0.5$\\
        Top-$K$ pruning selection & $0.4\times B$\\
        Pruning objective calculated with $P$ DDIM inversion steps & $0.5 \times T$ \\
        \bottomrule
    \end{tabular}
\end{table}

All experiments were run on single NVIDIA\texttrademark\ L40s\ or NVIDIA\texttrademark\ A100\ GPUs.

\section{Pseudo-code and Hyperparameters for Video Generation}
\label{app:sec:video-code}

We modify CogVideoX pipeline provided in Huggingface: \url{https://github.com/huggingface/diffusers/blob/v0.35.1/src/diffusers/pipelines/cogvideo/pipeline_cogvideox.py}.

We keep the logic same as images.

\begin{table}[h]
    \centering
    \caption{Sampling setup hyperparameters for long-video generation experiments}
    \label{tab:video_sampling}
    \begin{tabular}{lc}
        \toprule
        \textbf{Hyperparameter} & \textbf{Value} \\
        \midrule
        Denoising timesteps $T$ & $30$ \\
        Batch size $B$ & $10$ \\
        Composition weights $\gamma_{1:H}$ & $0.5$ \\
        Resampling schedule $U(t)$ & $\frac{T - t + 1}{T}(U_{T})$\\
        Maximum resampling steps $U_T$ & $10$\\
        Exploration ends at $k_e$ & $0.3$\\
        Pruning ends at $k_p$ & $0.6$\\
        Top-$K$ pruning selection & $0.4\times B$\\
        Pruning objective calculated with $P$ DDIM inversion steps & $0.5 \times T$ \\
        \bottomrule
    \end{tabular}
\end{table}

All experiments were run on single NVIDIA\texttrademark\ H100\ GPUs. More details about image and video generation code can be found here: \url{https://github.com/UtkarshMishra04/CDGS_imgvideo}.

\newpage
\begin{rebuttal}
\section{Scaling analysis: NFE and Wall clock times}

In general, we consider $T$ denoising iterations for which we perform $U$ steps of iterative resampling to get the candidate global plans and then perform $T$ steps of DDIM inversion steps to prune infeasible candidates. Eventually, at each denoising step, CDGS selects the best-K denoising paths. We repeat the selected denoising paths to fill up the batch for the next denoising step. Since we use stochasticity in the main denoising loop, the same denoising paths can lead to different clean samples.

Thus, we can compute the NFEs as:
\begin{equation}
    NFE = T \times U + T\times T
\end{equation}
If we consider model inference complexity to be $O(1)$, the computational complexity of CDGS is $O(T^2)$ if $T\geq U$ else it is $O(U^2)$. To give a comparison CDGS is $(U+T)$ times more expensive to run than na\"ive compositional sampling. 

To reduce the complexity and compute requirements, we perform some engineering-modifications: 
\begin{enumerate}
    \item we observe that early pruning does not help a lot since Tweedie estimates for noisy samples at higher noise levels are not very accurate, hence:
    \begin{enumerate}
        \item instead of always performing $U$ resampling steps, we gradually increase $U$ throughout the denoising process such that we do not overfit to bad denoising paths at earlier timesteps
        \item we can deploy pruning only for the last 20\% denoising iterations 
    \end{enumerate}
    this makes effective number of resampling steps approximately $U/2$.
    \item we also observe that abrupt direction and magnitude changes of score functions are more prominent in the initial DDIM inversion steps (eventually it stabilizes as noisy latents come in-distribution), allowing us to stop DDIM inversion steps at $T/2$. 
\end{enumerate}
This allows making CDGS only $(0.5U+0.1T)$ more expensive than na\"ive compositional diffusion. To give a practical example, by incorporating jit compilation, a single model inference for Stable Diffusion 2.1 takes $1.5$ secs on a NVIDIA\texttrademark\ L40s\ GPU and with $T=50$ it takes $75$ secs to generate panoramic image using na\"ive compositional sampling. With $U=10$ and pruning happening for $0.2T$ steps, with CDGS it takes around $700$ secs.

Compute and wall-clock time are completely dependent on the base local generative model and the number of inference steps required to generate a good sample from it. For example we observe that for toy and robotics domains, $T=50$ is sufficient to sample good solutions. Also, note that, for a batch of $B$ candidate global plan for horizon $H$ each with $M$ local segments, we construct a batch of local segments of size $B\times M$ to denoise all the local segments in parallel for every denoising step. This step depends on the available GPU memory, which limits the maximum batch size.

\subsection{Toy domain}
We analyze the runtime and success of scaling inference-time compute in the toy-domain. All experiments in this section were run on single NVIDIA\texttrademark\ V100\ GPUs.

\begin{figure}[t]
\centering
    \includegraphics[width=\linewidth]{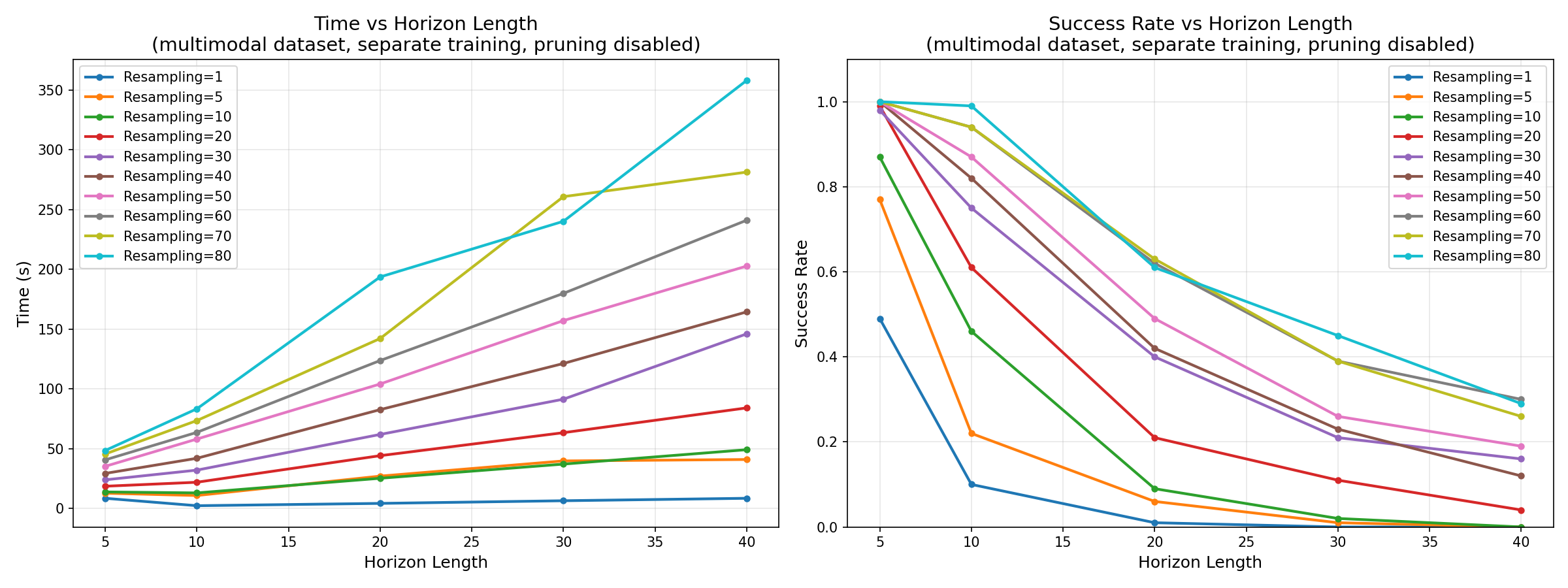}
    \caption{Left: runtime scales linearly with horizon length and resampling steps. Right: success rates decline over horizon lengths, but this is alleviated by additional resampling.}
    \label{fig:time_success_toy_domain}
\end{figure}

In our first experiment, we disable pruning and ablate the number of resampling steps $U$ as shown in~\autoref{fig:time_success_toy_domain}. We find that:
\begin{enumerate}
    \item wall-clock time scales linearly with the additional compute
    \item increasing resampling steps can address declining performance as horizons increase
    \item overall, a key finding is that the improvement in performance with increasing $U$ diminishes as the horizon increases
\end{enumerate}

\begin{figure}[t]
    \centering
    \includegraphics[width=\linewidth]{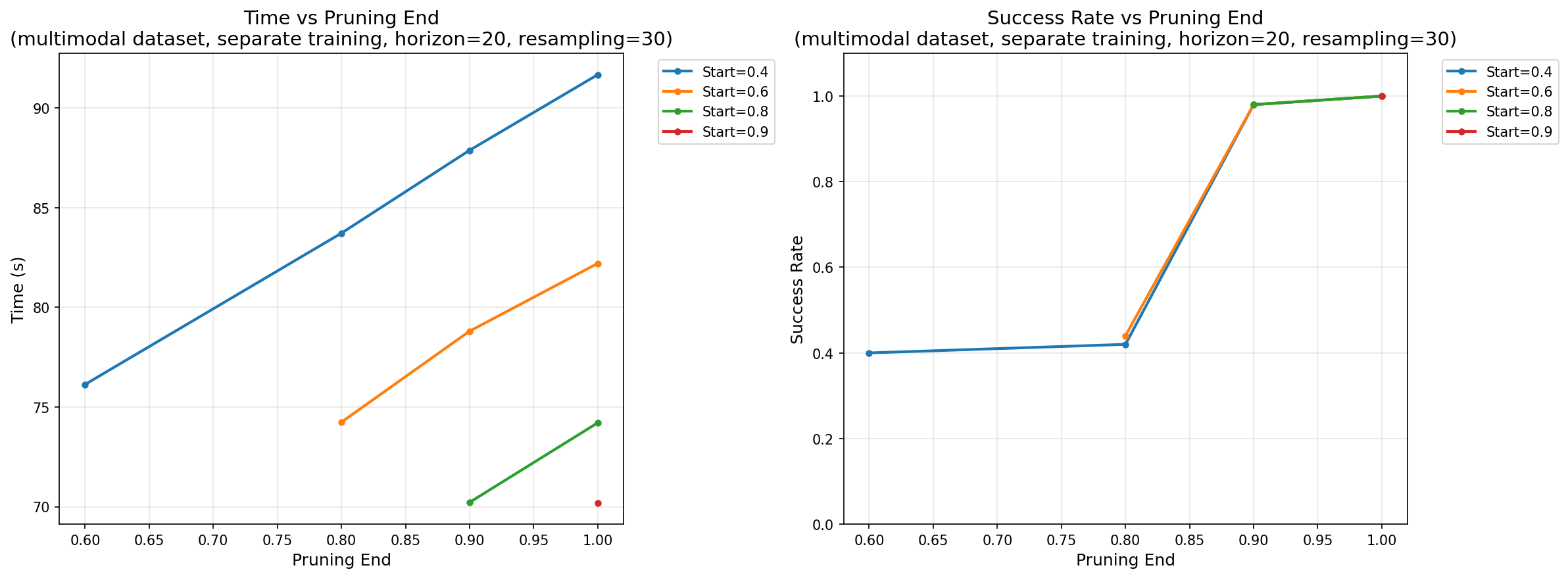}
    \caption{Comparison of time and success heatmaps for pruning}
    \label{fig:heatmaps_comparison}
\end{figure}


In our second experiment, we ablate the choice of the parameters for pruning: start and end. We find that:
\begin{enumerate}
    \item the cost of pruning increases wall-clock linearly.
    \item we perform this experiment with horizon $H=20$ and number of resampling steps $U=30$. Bu adding pruning, we note that with minimal increase in wall-clock time (around 5\%), we can push the success rate to be 100\%.
    \item One additional insight we obtained is that pruning until the end of the denoising process is essential. This supports our key insight that as Tweedie estimates get accurate at lower noise levels, pruning becomes more effective in selecting better denoising paths.
\end{enumerate}

It is worth noting that because of independently sampling local segments, the complexity of the problem increases exponentially with horizon. For example, for a horizon of $H=5$ and each transition having two feasible modes, there can be $2^H$ possible sequences of feasible factor modes; only two of them will be valid for coherent global plan synthesis. \textbf{CDGS is able to navigate this exponentially increasing domain by linearly scaling the compute and memory requirements.}

\clearpage

\subsection{OGbench domain}

We report the wall clock times and associated gain in performance for the OGbench Maze domains (similar for both PointMaze and AntMaze) in ~\autoref{fig:ogb_time_plot}. It is worth noting that CDGS uses more compute to scale performance even with na\"ive compositional methods.

\begin{figure}[h]
    \centering
    \includegraphics[width=\linewidth]{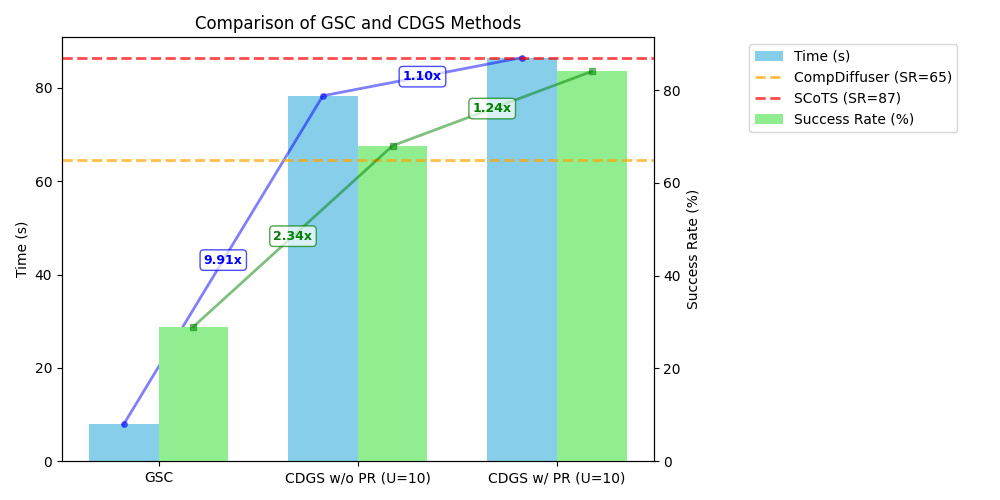}
    \caption{We show results for OGBench maze tasks. We observe that performance improves with adding resampling steps along with additional computational time. With pruning, we see more improvement in performance with only an additional 10\% compute time. Overall, CDGS with resampling and pruning takes around 10-12x more time than GSC. This relationship validates that CDGS scales linearly with number of resampling steps and pruning.}
    \label{fig:ogb_time_plot}
\end{figure}

It should be noted that CDGS with resampling and pruning can scale the performance of na\"ive compositional sampling, in a training-free manner, to an extent that:
\begin{enumerate}
    \item beats baselines like CompDiffuser~\cite{luo2025generative} that use overlap information while training and learn an overlap conditioned score function.
    \item performs on par with baselines like SCoTS~\cite{lee2025state} that use data augmentation to synthesize datasets with long-horizon data and train a policy on the new dataset.
\end{enumerate}

\end{rebuttal}

\begin{rebuttal}

\section{Compositional Diffusion with Guided Search: Complete Algorithm for Motion Planning}
\label{app:algorithm_motionplan}

\textbf{Problem with the TAMP benchmark:} TAMP benchmark is based on skill-level action learning. For example, for \texttt{push} skill, this means that instead of learning the low-level end-effector motion, we are learning start pose of the end effector wrt the target object's position (here, cube) and by how much we want to gripper to move to complete the skill execution. This structure implies that for every skill, only certain objects can move in the environment, and the other objects remain static.

\textbf{Subproblem 1: What happens when more objects move in the predicted state of the planner?} Since the planner is trained on diverse set of skill transitions and predicts the sequence of $\{(s_i, \pi_i, a_i, s_{i+1})\}$, it is likely that for a particular predicted skill $\pi_i$, for example \texttt{pull}, objects other than the target cube move in the predicted next state of the transition. For DDIM inversion objective, even if the planned transition of the target object is correct, it will reject the transition as other objects have  moved too.

\textbf{Solution:} We use learned forward dynamics model per skill to ensure that only the objects relevant to the predicted skill move for a planned transition. Basically for every predicted skill in the planned sequence of CDGS $\{(s_i, \pi_i, a_i, s_{i+1})\}$, we use forward dynamics model $f_{\pi_i}$ to overwrite $s_{i+1} = f_{\pi_i}(s_i, a_i)$ such that only the pose of target objects (hook, gripper and target cube in case of \texttt{pull} skill) to change and other objects remain static. This allows DDIM inversion to evaluate and score planned local transitions appropriately. 

\textbf{Changes in algorithm to incorporate the solution:}

\begin{algorithm}[H]
\caption{\ours}
\label{algo:search_algo_tamp}
\begin{algorithmic}[1]
\Require Start $x_{s}$, Goal $x_{g}$, Planning horizon $H$
\Require Diffusion noise schedule, 
\Require Pretrained local plan score function $\epsilon_\theta(y^{(t)},t)$, 
\Require number of candidate plans $B$, number of elite plans $K$ at every step
\State Initialize $B$ global plan candidates: $\tau^{(T)}$
\State $\tau^{(T)} = (y^{(T)}_1 \circ \dots \circ y^{(T)}_M) \sim \mathcal{N}(0, \textbf{I})$
\For{$t = T, \dots, 1$}
    \State $\epsilon(\tau^{(t)},t)$ = ComposedScore($\tau^{(t)}, t, \epsilon_\theta, x_{s}, x_{g}$)
    \State $\hat{\tau}_0^{(t)} = (\tau^{(t)} - \sqrt{1-\alpha_t} \epsilon(\tau^{(t)},t))/\sqrt{\alpha_t}$
    \State \changes{$\hat{\tau}_{0,new}^{(t)} = \text{LearnedForwardDynamics}(\hat{\tau}_0^{(t)}) $}
    \State \changes{Rank plans using $J(\hat{\tau}_{0,new}^{(t)})$ ~\autoref{eq:pruning_objective}}
    \State Select best-$K$ global plans
    \State Repopulate candidates using filtered plans
    \State $\tau^{(t-1)} \sim p(\tau^{(t-1)}|\tau^{(t)}, \hat{\tau}_0^{(t)})$ ~\autoref{eq:tweedie}
\EndFor
\State return $\tau^{(0)}$
\end{algorithmic}
\end{algorithm}

\end{rebuttal}

\end{document}